\documentclass[conference]{IEEEtran}
\usepackage{times}

\usepackage[numbers]{natbib}
\usepackage{multicol} 
\usepackage[dvipsnames]{xcolor}
\usepackage[bookmarks=true]{hyperref}
\usepackage{graphicx}
\usepackage{amsmath} 
\usepackage{amssymb} 
\usepackage{bm}
\usepackage{amsmath}
\usepackage{algpseudocode}
\usepackage{algorithm}
\usepackage{todonotes}
\usepackage{bbm}
\usepackage{caption}
\usepackage{subcaption}
\newcommand{\apf}{{\sc apf}}
\newcommand{\ba}{\mathbf{a}}
\newcommand{\as}{(a_1, \ldots, a_n)}
\newcommand{\feas}{\phi}
\newcommand{\Feas}{\Phi}
\newcommand{\Data}{\mathcal{D}}

\newcommand{\argmax}[1]{\underset{#1}{\text{argmax}\;}}
\newcommand{\A}{\mathbf{A}}
\newcommand{\info}{\mathrm{I}}
\newcommand{\ent}{\mathrm{H}}
\newcommand{\expect}[1]{\mathbb{E}_{#1}}
\newcommand{\bald}{{\sc bald}}
\newcommand{\complete}{{\it complete}}
\newcommand{\sequential}{{\it sequential}}
\newcommand{\incremental}{{\it incremental}}
\newcommand{\greedy}{{\it greedy}}
\newcommand{\tgn}{{\sc tgn}}

\let\svthefootnote\thefootnote
\newcommand\freefootnote[1]{%
  \let\thefootnote\relax%
  \footnotetext{#1}%
  \let\thefootnote\svthefootnote%
}

\begin{document}

\title{Active Learning of Abstract Plan Feasibility}

% You will get a Paper-ID when submitting a pdf file to the conference system
% \author{Author Names Omitted for Anonymous Review. Paper-ID [add your ID here]}
\author{
\IEEEauthorblockN{Michael Noseworthy*, Caris Moses*, Isaiah Brand*, Sebastian Castro, \\
Leslie Kaelbling, Tomás Lozano-Pérez, Nicholas Roy} 
\IEEEauthorblockA{MIT, CSAIL}
}

%\author{\authorblockN{Michael Shell}
%\authorblockA{School of Electrical and\\Computer Engineering\\
%Georgia Institute of Technology\\
%Atlanta, Georgia 30332--0250\\
%Email: mshell@ece.gatech.edu}
%\and
%\authorblockA{Twentieth Century Fox\\
%\and
%\authorblockN{James Kirk\\ and Montgomery Scott}
%\authorblockA{Starfleet Academy\\
%San Francisco, California 96678-2391\\

% avoiding spaces at the end of the author lines is not a problem with
% conference papers because we don't use \thanks or \IEEEmembership

%Homer Simpson\authorrefmark{2},
%Montgomery Scott\authorrefmark{3} and
%Eldon Tyrell\authorrefmark{4}}
%Telephone: (800) 555--1212, Fax: (888) 555--1212}
%\authorblockA{\authorrefmark{4}Tyrell Inc., 123 Replicant Street, Los Angeles, California 90210--4321}}

\maketitle

% NOTE: included for page numbering purposes as per 
% https://tex.stackexchange.com/questions/52729/forcing-page-numbers-with-ieeetran
% this should be removed eventually
% \thispagestyle{plain}
% \pagestyle{plain}

\begin{abstract}
Long horizon sequential manipulation tasks are effectively addressed hierarchically:
at a high level of abstraction the planner searches over abstract action sequences, and when a plan is found, lower level motion plans are generated. 
Such a strategy hinges on the ability to reliably predict that a feasible low level plan will be found which satisfies the abstract plan. 
However, computing \emph{Abstract Plan Feasibility} (\apf{}) is difficult because the outcome of a plan depends on real-world phenomena that are difficult to model, such as noise in estimation and execution. 
In this work, we present an active learning approach to efficiently acquire an \apf{} predictor through task-independent, curious exploration on a robot. 
The robot identifies plans whose outcomes would be informative about \apf{}, executes those plans, and learns from their successes or failures.
Critically, we leverage an \emph{infeasible subsequence property} to prune candidate plans in the active learning strategy, allowing our system to learn from less data.
% Critically, we leverage the in-progress \apf{} predictor to prune candidate plans in an information-theoretic active-learning framework, allowing our system to learn from less data than is used in the current state-of-the-art model-learning approaches.
% By carefully designing a model class with appropriate inductive biases and a representation of epistemic uncertainty, as well as constraining the space from which candidate plans are sampled, the robot can quickly find informative abstract plans to execute. 
We evaluate our strategy in simulation and on a real Franka Emika Panda robot with integrated perception, experimentation, planning, and execution.
In a stacking domain where objects have non-uniform mass distributions, we show that our system permits real robot learning of an \apf{} model in four hundred self-supervised interactions, and that our learned model can be used effectively in multiple downstream tasks% (e.g., constructing the tallest tower or tower with the longest overhang).
\footnote{An accompanying video can be found at {\color{blue} \url{https://youtu.be/UF-SjGm20Mw}}}.
\end{abstract}

\IEEEpeerreviewmaketitle

%\section{Introduction}
%\input{sections/intro.tex}

%\section{Things we could emphasize}
%\input{sections/ideas.tex}

%\clearpage
\section{Introduction}
\label{sec:intro}
\freefootnote{*Equal contribution.}

Long horizon sequential manipulation tasks still pose a challenging problem for robotic systems. 
Tasks such as assembly depend on using many objects with varying physical properties. 
Finding a plan to achieve a task in these domains consists of reasoning over large spaces that include discrete action plans, as well as low level continuous motion plans.
%and object arrangements.

% Motivate high-level framework in which we are working.
These problems can be effectively addressed hierarchically: at the highest level of abstraction the system searches over plausible \emph{abstract} action sequences, and at the lower level it plans for detailed \emph{concrete} motion plans and object interactions.
The complexity of the search space for the concrete planner is greatly reduced when constrained by the abstract action sequence.
Further computational efficiencies can be gained if we lazily \cite{haghtalab2017provable} postpone concrete planning until we have a complete abstract plan that is likely to succeed, avoiding the need to query the concrete planner multiple times. A version of this approach is used in {\em skeleton-based} task and motion planning systems \cite{lozano-perez_constraint-based_2014, driess2020probabilisticTAMP, kim2020learning}.
%
%NR: I deleted the following sentence because the concrete plan isn't needed to ensure correctness, it's needed to convert abstract plan into something that can be executed. 
%In general, to be sure of correctness it is necessary to find a complete concrete plan using detailed models of the robot's interaction with the physical world. 
%This is expensive in terms of computation, and learning accurate physical models is expensive in terms of training examples.
%Nick suggestion to replace the above sentence -- Subject to constraints imposed by the high level abstract plan. The high level abstract plan constrains the search space of the low level planner allowing the low level planner to find plans much more efficiently than it would have in the original problem. 
% 
%\nrnote{However, the computational efficiency induced by first finding an abstract plan may also lead to infeasible plans at the lower level. The abstract planner may need to query the lower level plan multiple times in order to ensure a feasible low level plan does exist, reducing the computational savings induced by the abstract planner.}

% Introduce the APF problem.
%Accordingly, we generally wish to lazily \cite{haghtalab2017provable} postpone concrete planning until we have a complete abstract plan that we believe is likely to succeed.
The success of this lazy strategy hinges on our ability to predict whether an abstract action sequence will be feasible to execute.
In this work, we call an abstract action sequence feasible if both the concrete planner returns a solution and this solution is reliably executed in the real world with the intended outcome.
If the abstract planner does not take into account errors in execution due to phenomena unmodeled by the concrete planner, the robot may end up attempting a plan that fails during execution.
% We aim to predict feasibility so that we only do concrete planning when we are confident the resulting plan will be feasible.
%\nrnote{Being able to perfectly identify whether an abstract plan sequence will succeed without computing the corresponding concrete plan is in general not possible --- the computational savings of an abstraction typically leads to some loss relative to the concrete plan.} 

Fortunately, even an approximately correct estimator of {\em abstract plan feasibility} (\apf) can offer huge computational advantages during planning.  
%A version of this approach is used to good effect in {\em skeleton-based} task and motion planning systems \cite{lozano-perez_constraint-based_2014, driess2020probabilisticTAMP, kim2020learning}.
In some cases it may be possible to approximate \apf{} via coarse-grained simulation. However, this strategy still requires a coarse dynamics model, which may not capture complex phenomena needed to accurately predict feasibility in the real world.

% Difficulty of modeling APF/motivation for learning it.
% Although in some cases it may be possible to approximate \apf{} via coarse-grained simulation, true feasibility will generally depend both on low-level planning details and on the accuracy of the domain model. 
% In manipulation tasks such as assembling objects, checking for physical stability of assemblies using a physical model may require considering complex real-world phenomena such as the robot's motor capabilities, errors in perception, and properties of objects.
%(including geometric, dynamic, and material properties). % that dictate how they interact with one another.
% Furthermore, it is difficult to know {\em a priori} the extent to which these phenomena affect \apf{} for each specific domain, and therefore to which level of detail we should account for them in our models.
% NOTE: i took out most of the above paragraph except for mentioning coarse-simulation (below). My reasoning was that we already talk about the difficulty of the domain, and the difficulty of acquiring a model earlier in the introduction. open to push-back tho

% Difficulties that arise if we want to learn the model.
%In some cases it may be possible to approximate \apf{} via coarse-grained simulation. However, this strategy still requires a coarse dynamics model, which may not capture complex phenomena needed to accurately predict feasibility in the real world. 
Instead, we explore a strategy in which we learn a model that predicts \apf{} by exploring the space of real plan executions without a specific planning problem or task at hand --- a form of curious exploration \cite{oudeyer2005playground}.
Data efficiency is a primary concern in enabling real robot learning of feasibility models.
Here, a training instance is the execution trace of an abstract plan, labeled by success or failure. Labeling each such plan is very expensive as it involves finding and executing a concrete motion plan, potentially taking several minutes on a real robot.
Furthermore, due to the combinatorial input space of abstract action plans, randomly executing actions is unlikely to elicit interesting behavior.

\begin{figure}
    \centering
    \includegraphics[clip, trim={0 0cm 0 0cm}, width=0.85\columnwidth]{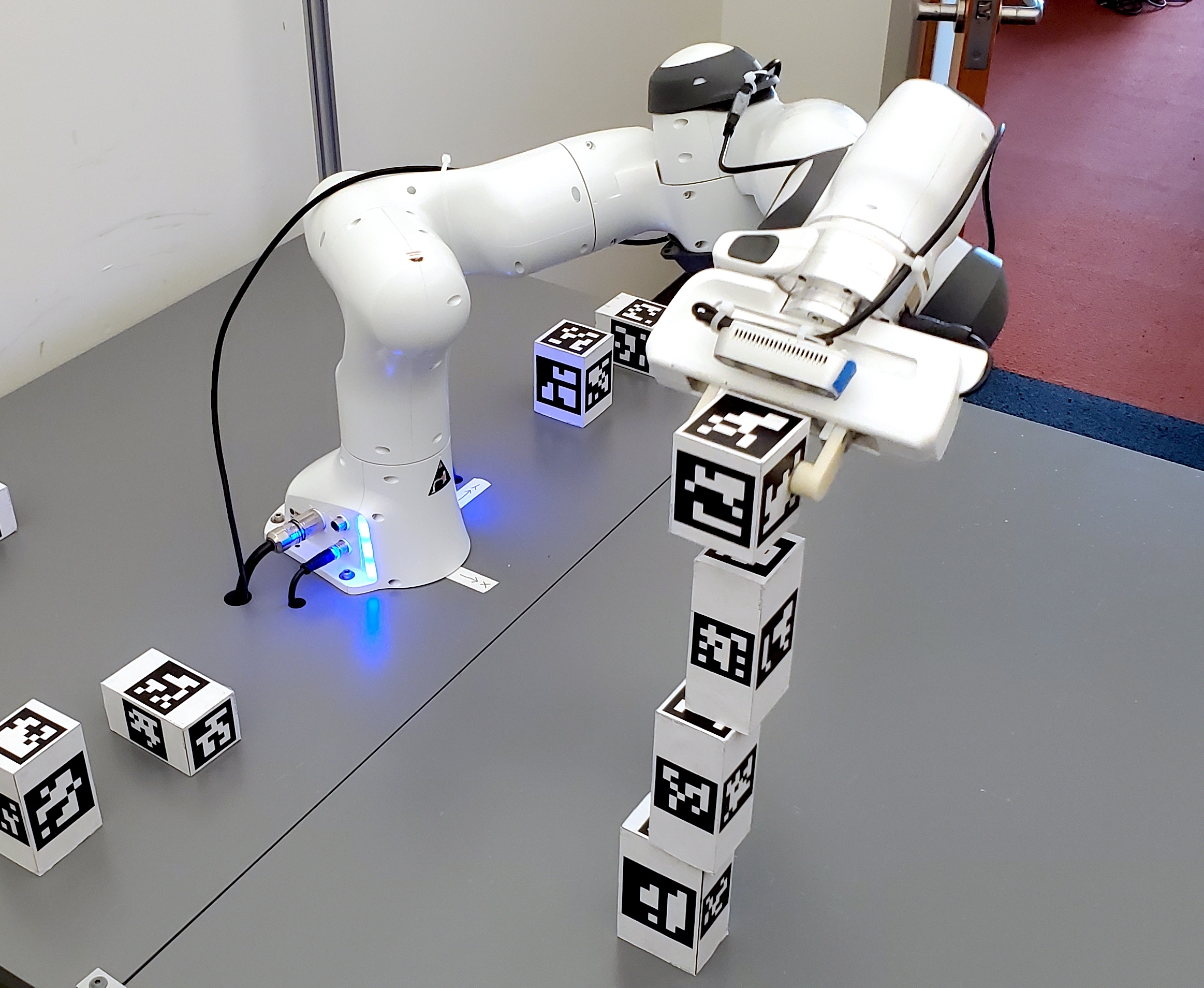}
    \caption{The Franka Emika Panda robot constructing a tower to improve its understanding of plan feasibility. A wrist-mounted camera refines object pose estimates for precise grasping.}
    \label{fig:panda_intro}
    \vspace{-12pt}
\end{figure}

\begin{figure*}[ht]
\begin{center}
\includegraphics[width=0.9\linewidth]{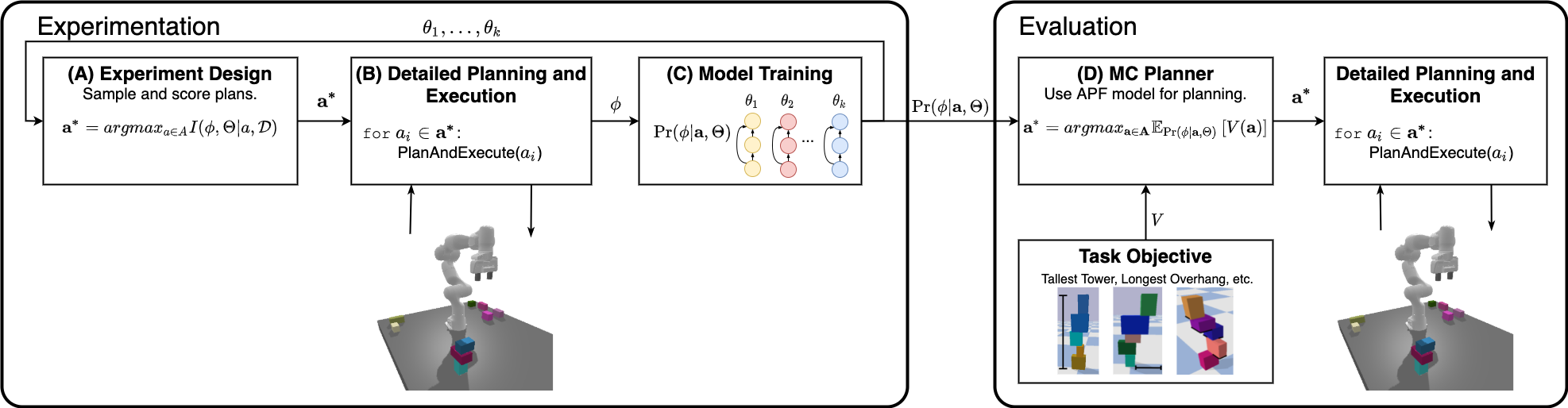}
\end{center}
\caption{\label{fig:method_overview} The proposed system for learning \emph{Abstract Plan Feasibility} (\apf{}) operates in two phases. \textbf{Experimentation Phase} (left) The robot iteratively designs and executes experiments that improve its \apf{} model. \textbf{(A)} Using its current model, the robot selects the abstract action sequence, $\mathbf{a^*}$, that minimizes its entropy over the \apf{} model. \textbf{(B)} The robot then computes and executes a concrete motion plan for $\mathbf{a^*}$. \textbf{(C)} After observing the true plan feasibility, $\phi$, the robot uses this new labeled data to update its \apf{} model, represented as an ensemble of neural networks. \textbf{Evaluation Phase} (right) Once an \apf{} model has been learned, the robot can use it to perform various tasks, such as building the tower with the longest overhang from a given set of blocks.}
\vspace{-10pt}
\end{figure*}

% Motivation for Active Learning.
% Note (Mike): I added a modified version of the middle part back in. I think it does a good job making clear the input space for the active learner is plans.
To address the data efficiency problem, we observe that in the process of training the \apf{} model, some observations may be more valuable than others. 
Active learning is a technique for identifying unlabeled instances that are most informative in learning a target concept. %, driving down the data complexity required to achieve good performance from the learner. 
% (Note): Mike - readding this. I think it is a good story about what the robot does. Maybe not for space though...
%A robot that could identify plans whose outcomes would be informative about \apf{}, execute those plans and learn from the subsequent success or failure, may be able to learn with much less data. % than is used in the current state-of-the-art model-learning approaches. 
The technical challenge is how to find plans of interest --- an active learning approach requires both a way to generate candidate plans, and a way to score how informative a candidate plan might be given the current model.

% Our Active Learning solution. 
% Note: (Mike) I currently put this paragraph to answer the two question posed at the end of the previous paragraph. Open to other suggestions.
To determine how informative a plan is with respect to the learned \apf{} model, we adopt an information-theoretic active learning approach \cite{mackay1992objective, houlsby_bayesian_2011}.
To generate candidate plans, we exploit an important property of abstract action sequences: for an action sequence $\as$, if any prefix $(a_1, \ldots, a_i)$ is infeasible, then any longer prefix $(a_1, \ldots, a_j)$ for $i <j \leq n$ is also infeasible. This \emph{infeasible subsequence property} gives us leverage %not only in planning but also 
during data acquisition. A complex plan instance may contain many elements that are highly informative for model learning, but will never be experienced because early elements in the plan will fail with high probability.

% \nrnote{By evaluating the information gain of each step of the plan, we ensure that the plan is highly informative even when the plan fails in the first few steps.}
%This requires a model class capable of representing uncertainty of our \apf{} predictions. In this work we represent this using an ensemble of neural networks. % \inote{we do talk about about this in the Abstract Plan Feasibility Model section, so it could be removed here}
%Comment from Nick -- Given that we talk about aleatoric and epistemic uncertainty a couple of times, I think we should define both and explain their importance here. Part of the reason we want to learn the APF is because of aleatoric uncertainty, but we need to collect enough data that we aren't subject to epistemic uncertainty. And we need to collect the right data so we aren't confusing the two.
% Caris response: I agree, but I think the discussion should be in the active learning section, not the intro.
% Experiments/real robot results.
We apply this active learning strategy to the concrete problem of stacking blocks with a real robot, where the blocks are each unique and have non-uniform mass distributions. 
The robot autonomously designs, plans, and executes experiments to learn a feasibility model using a Franka Emika Panda robot arm (Figure~\ref{fig:panda_intro}). The robot is also capable of resetting the world state after each experiment, enabling continuous autonomous experimentation.
The learned feasibility predictor is later used to build towers with previously unseen blocks that satisfy several different objective functions, including the tallest possible tower or the tower with the longest overhang.
This sample-efficient autonomous learning process relieves engineers from supervising data collection, resetting the experimental environment, and having to specify accurate dynamics models for planning. This results in a highly flexible and robust system for planning and executing complex action sequences in the real world.

%In summary, our contributions that enable efficient learning of \emph{Abstract Plan Feasibility} are:
%\begin{itemize}
%    \item An active learning approach to select test plans that are as informative as possible; 
%    \item A novel method to leverage abstract plan feasibility in the synthesis of informative plans;
%    \item A robotic system which conducts autonomous self-supervised learning via integrated perception, experimentation, planning, and execution.
%\end{itemize}

In summary, our contributions are: 
\begin{itemize}
    \item A method to learn an {\em Abstract Plan Feasibility} model by synthesizing hypothetical plans; 
    \item A data acquisition approach which leverages the {\em infeasible subsequence property} when sampling potential plans;
    \item A robotic system which conducts autonomous self-supervised learning via integrated perception, experimentation, planning, and execution.
\end{itemize}

\section{Problem Formulation}
\label{sec:problem}

\label{sec:exp-design}
Our objective is to learn a model that predicts the success of an abstract action sequence when it is executed by the robot.  That is, to learn the parameters $\Theta$ that predict
\[
\Pr(\phi \mid \ba; \Theta),
\label{eq:apf}
\]
where $\phi \in \{0, 1\}$ is the success of the sequence of abstract actions, $\ba = \as$. Furthermore, we wish to learn $\Theta$ using as few labeled action sequences $(\ba, \phi)$ as possible.
%The feasibility probability distribution represents uncertainty over our predictions. 
%This uncertainty could be due to the fact that we have not yet learned the correct parameter values, referred to as {\em epistemic uncertainty}. It could also come from randomness in the underlying process being modeled, referred to as {\em aleatoric uncertainty}. The objective of the active learning framework is to collect data that allows us to disambiguate between these two types of uncertainty such that we can reduce {\em epistemic uncertainty} and make accurate predictions in spite of systemic {\em aleatoric uncertainty}.
%Once the \apf{} model is learned, it should represent {\em aleatoric uncertainty} due to unmodeled behaviors of the detailed planning process and execution of actions in the world.

%Note that in general, \apf{} may need to consider the initial state in which a plan is to be executed, however in this work we make the assumption that the initial state is always the same, so that plan feasibility is simply a function of the sequence of actions in the plan. The method applies regardless of whether initial state is fixed.
%Note that in general, \apf{} may need to consider the initial state in which a plan is to be executed, however in this work we make the assumption that all abstract actions are feasible first actions in a sequence, and do not depend on the initial state. 
Note that in general, \apf{} may need to consider the initial state in which a plan is to be executed, however in this work we make the assumption that the first abstract action in a sequence will be feasible, and does not depend on the initial state. 
The method applies regardless of whether this assumption is made.

To learn this \apf{} model, our system operates in two phases as illustrated in Figure~\ref{fig:method_overview}.
In the {\em experimentation phase}, the robot curiously explores the space of possible plans, learning the parameters of the \apf{} model;  in the {\em evaluation phase}, the robot is given specific goals to achieve, and uses the learned \apf{} model to efficiently plan over abstract action sequences.
% Our system, which learns this \apf{} model, operates in two separate phases as illustrated in figure~\ref{fig:method_overview}.
% In the {\em experimentation phase}, the robot ``curiously'' plays with the objects available to it, learning the parameters of the \apf{} model;  in the {\em execution phase}, the robot is given novel objects and goals to achieve, and it uses the learned \apf{} model to efficiently plan successful action sequences.  

Both phases depend on the ability to construct and execute concrete plans on a real robot given an abstract action sequence.
% We assume that a detailed planning and execution system is available that can integrate perception, planning, and control to carry out an abstract action sequence and observe whether or not it was successful.
We assume that a system is available that can integrate perception, planning, and control to execute an abstract action sequence and observe whether or not it was successful.
In Section~\ref{sec:implementation} we describe our implementation of this system in detail.

% In the following, we assume the existence of a procedure \textproc{detailedPlanEx}, which takes a sequence of abstract actions $\ba$ and the current robot state, and performs closed-loop perception, planning, and control to execute actions $a \in \ba$ in sequence until it determines that $\ba$ has been successfully carried out or has become impossible.  It returns a Boolean value indicating whether it succeeded.

\subsection{Experimentation phase}

During the experimentation phase, the robot performs active learning to efficiently design, plan, and execute abstract action sequences that are informative about \apf{}.
This phase operates in a loop.
At each iteration, the robot first generates candidate action sequences that may be informative according to some \emph{sampling strategy}.
Each action sequence is then scored using the current model according to some acquisition function, $f: \A \rightarrow \mathbb{R}$.
The highest scoring plans are executed on the robot to obtain feasibility labels which are then used to update the model for future iterations of the active learning loop.
The effectiveness of active learning in reducing data complexity depends critically upon the choice of acquisition function and sampling strategy --- or which action sequences are considered as possible experiments. 
These choices are discussed in more detail in Section~\ref{sec:active-learning}.

%The experimentation phase operates, as shown in algorithm~\ref{exper_pseudocode}, as follows.  It starts by initializing $n_m$ separate parameter sets $\theta$ for the {\sc gnn}s representing the $\apf$ model.  It then iterates $n_i$ through an experimentation and training loop.  It is parametrized by an action sampling strategy (one of \allseq{}, \sequential{}, or \incremental{}), which it uses to generate $n_s$ sample action sequences $A_hat$ (note that the \incremental{} strategy depends on the current world state and is really only sensible for $n_b = 1$).  Then, the approximate \bald{} objective is used to select the best batch of $n_b$ action sequences $A_batch$ from $A_hat$.  These action sequences are executed, yielding results $\phi_batch$, and the test sequences and results are added to the data set.  Finally, the parameters $\theta_i$ for each model in the ensemble are updated and a new iteration begins.

% 
% \begin{scriptsize}
% \begin{verbatim}
%     experiment(theta, sampler, scorer, world_state, n_i)
%         D = {}
%         for iteration in n_i
%             A_hat = sampler(A, L, D, n_s, world_state)
%             A = argmax(scorer(A_hat ; theta))
%             Phi_batch = detailedPlanEx(a, world_state) 
%             D = D u (A, Phi)
%             theta = Update(theta, D)
%         return Theta
% \end{verbatim}
% \end{scriptsize}

\subsection{Evaluation phase}

After the \apf{} model has been learned, it can be used to achieve multiple objectives. 
We assume that the robot is given an objective function $V$ which maps action sequences to real values and its goal is to find the abstract action sequence, $\ba \in \A$ with the highest expected value,
%\[\ba^*= \argm[x{\{\ba \in \A \mid \Pr(\feas \mid \ba) > 1-\epsilon\}} V(\ba)\;\;,\]
\begin{equation}
    \ba^* = \argmax{\ba \in \A} \mathbb{E}_{\Pr(\phi|\ba, \Theta)}\left[ V(\ba) \right]
\end{equation}
where the expectation takes into account the likelihood that an abstract action sequence is  feasible when executed on the real robot.
To maximize this objective, we use a Monte Carlo planner which randomly samples action sequences and selects the one with the maximum expected value.

\section{Active Learning of Abstract Plan Feasibility}
\label{sec:active-learning}
Collecting data on real robot platforms is both time and cost-intensive.
% In addition, when considering the space of sequential actions with multiple object interactions, the search space is prohibitively large. \inote{search space over what? will fix in a sec}
To minimize the amount of data needed to learn the \apf{} model, we take an information theoretic approach to active learning \cite{houlsby_bayesian_2011}.
Concretely, we maintain a distribution over model parameters that are consistent with the data we have observed so far.
Our objective is to select new data that minimize the entropy over this distribution as quickly as possible.
Efficient active learning requires: (1) a model class that captures uncertainty in model parameters, (2) a way to score unlabeled plans based on how informative they may be, and (3) a method of generating potentially informative plans.
We discuss each of these in turn.

\subsection{Abstract plan feasibility model}
\label{sec:method-apf-model}

Our \apf{} model, $\Pr(\feas \mid \ba; \Theta)$, aims to capture the uncertainty in the underlying stochastic process of predicting the feasibility of abstract action sequences.
This uncertainty can be attributed to phenomena such as the robot's motor capabilities, errors in perception, or unmodeled behaviors of the planning process, and is referred to as {\em aleatoric uncertainty}. 
Our goal is to learn parameters $\Theta$ such that this uncertainty is adequately captured and our model can be leveraged, along with a low level planner, to achieve a goal.

We take a Bayesian approach to learning the model parameters, and maintain a distribution over the parameter space, $\Pr(\Theta)$.
This distribution aims to capture the uncertainty we have regarding the accuracy of our predictions, referred to as {\em epistemic uncertainty}. 
In general, for complex model classes such as neural network classifiers, an explicit representation of $\Pr(\Theta \mid \Data)$ for training data $\Data$ is difficult to construct or update with new data.
We therefore follow the strategy of \citet{beluch_power_2018} and represent this uncertainty with an ensemble of $N$ models, $(\theta_1, \ldots, \theta_N)$, where ${\theta_i} \in \mathbb{R}^{d}$.
Initial parameters are drawn independently at random and are updated to incorporate new data via gradient descent.

%Because our objective is to learn the \apf{} model from as few data samples as possible, it is critical that we make our {\em epistemic uncertainty} about the model explicit.  We represent our particular model using a parameter vector $\theta$ in the space of possible \apf{} models $\Theta$, and the prediction of a particular model is $\Pr(\feas \mid \ba; \theta)$. If we were to maintain a distribution on the model parameter space, $\Pr(\Theta)$, the entropy of that distribution would be the {\em epistemic uncertainty} of our model.

%%At the same time, we desire a model class that is flexible enough to learn about plan feasibility in the real world. 
%However, in general, for complex model classes such as our proposed neural network classifier, an explicit representation of $\Pr(\Theta \mid \Data)$ for training data $\Data$ is too difficult to construct or update with new data.  We therefore follow the strategy of \cite{beluch_power_2018} and represent this uncertainty using an ensemble of $N$ models, parameterized by $(\theta_1, \ldots, \theta_N)$. Initial parameters are drawn independently at random and are updated to incorporate new data via gradient descent.

%% Although domain-specific models could be applied, in this work we consider neural networks architectures to maintain generality. This decision highlights the data-efficiency of the \bald{} acquisition strategy; our model achieves acceptable performance on downstream tasks after collecting only 400 towers in the experimentation phase.

The design of the models in the ensemble is selected to match the underlying structure of the prediction problem. 
In a naive implementation, we can directly estimate feasibility from the entire sequence of actions. We will refer to this approach as a \complete{} model, denoted by $\Theta_{comp}$.
However, we observe that the \emph{infeasible sub-sequence property} provides a strong constraint on the set of feasible plans: sequence $\as$ is only feasible if all of its prefixes are also feasible. As such, the model can instead learn the probability a specific action is feasible given all previous actions were feasible. We will refer to a model that considers this property as $\Theta_{ss}$.
Under this model, the feasibility of a plan, $\ba$, is:
\begin{equation}
\label{eq:theta-ss}
    \Pr(\Phi_{1:n} \mid \ba, \Theta_{ss}) = \prod\limits_{i=2}^n \Pr(\Phi_i \mid a_{1:i}, \Phi_{1:i-1} = \bm1; \Theta_{ss}),
\end{equation}
where $\Phi_i$ represents whether action $a_i$ is feasible given $a_{1:i-1}$ were feasible, and we use the subscript $1:n$ to refer to a sequence of variables. In our method we assume initial actions are feasible, meaning $\Pr(\Phi_1 = 1 \mid a_1) = 1$.

The $\Theta_{comp}$ model only considers full action sequences, and therefore entire plans correspond to a single label once they are executed. On the other hand, $\Theta_{ss}$ requires labels after each action is executed.

In both of these models, the length of plans for which we require predictions varies.  Therefore, we use {\em graph neural networks} ({\sc gnn}s), which make predictions based on aggregations of local properties and relations among the input entities, and exploit parameter tying to model global properties of plans of arbitrary size using a fixed-dimensional parameterization $\Theta$.  For a detailed description of {\sc gnn}s, see the overview by \cite{zhou2018graph}.
More details about the specific architecture we use can be found in Section \ref{sec:imp-learning}.

%%We take advantage of the {\em infeasible sub-sequence} property of the \apf{} problem to design a special connectivity structure for our {\sc gnn} models.  As shown in figure~\ref{fig:graph_network}, the node corresponding to action $a_i$ is connected to each node representing $a_j$ for $1 \leq j < i$.  This encodes the structural property that the feasibility of step $i$ in the plan depends on all previous actions.  To make a prediction, an encoding of $\ba$ is fed into the network, some number of rounds of message passing is performed on the graph using the local neural network models $\theta$, and the average activation level of each node is computed and passed through a sigmoidal transformation to obtain a prediction $\Pr(\feas \mid \ba)$.
%\lpknote{We might want to put more details in here?}

\subsection{Entropy reduction}

Following \cite{mackay1992objective, cohn1996active}, we guide our active learning by picking a sequence of data $\Data$ that maximally reduces the entropy of $\Pr(\Theta \mid \Data)$. The general problem of designing a sequence of experiments to minimize entropy --- or equivalently, maximize information gain --- is a difficult sequential decision-making problem.  Fortunately, due to sub-modularity of the objective, a myopic approach that considers only the next experiment to conduct can be shown to be a good approximation to the optimal experimentation strategy \cite{Nemhauser1978submodular}. 

Given a model distribution $\Pr(\Theta \mid \Data)$ that depends on the data we have seen so far, $\Data$, we choose the action sequence $\ba \in \A$ that reduces the entropy of the posterior distribution as much as possible: 
% \inote{this seems to imply that we've already marginalized out $\phi$? We may need to have $\ent(\Theta \mid \Data, \ba, \phi)$. I just went back to looked at the original paper, and they just drop the form with the expectation over $\phi$ with no explanation...}
% \begin{equation}\label{eq:theta_entropy_objective2}
% \ba^* = \argmax{\ba\in\A} \ent(\Theta \mid \Data) - \ent(\Theta \mid \Data, \ba). 
% \end{equation}
\begin{equation}\label{eq:theta_entropy_objective2}
\ba^* = \argmax{\ba\in\A} \ent(\Theta \mid \Data) - \expect{\feas \sim \Pr(\cdot \mid \Data, \ba)} \left[ \ent(\Theta \mid \Data, \ba, \feas) \right]
\end{equation}
While the robot can select the plan, $\ba$, to experiment with, it cannot select the outcome $\feas$, so to compute this quantity, we have to take an expectation over the outcome, using our current model distribution.
% \[\ent(\Theta \mid \Data, \ba) = \expect{\feas \sim \Pr(\cdot \mid \Data, \ba)} \left[ \ent(\Theta \mid \Data, \ba, \feas) \right].\]
% where 
% \[\Pr(\feas \mid \Data, \ba) = \int_\theta \Pr(\feas \mid \ba; \theta) \Pr(\theta \mid \Data).\]

Estimating the entropy over a high-dimensional parameter space is expensive, so we follow the approach of~\citet{houlsby_bayesian_2011} to reformulate the objective in (\ref{eq:theta_entropy_objective2}) as: 
\begin{align}\label{eq:y_entropy_objective2}
\ba^* &= \argmax{\ba\in\A} \info(\Feas : \Theta \mid \Data, \ba) \\
&= \argmax{\ba\in\A} \ent(\Feas \mid \Data, \ba) - 
                          \expect{\Theta \sim \Pr(\cdot \mid \Data)} \left[ \ent(\Feas \mid \ba; \Theta) \right],
\end{align}
allowing the computation of entropies to take place in the lower-dimensional label space, $\Feas$. This is known as {\em Bayesian Active Learning by Disagreement}, or \bald{}.

The \bald{} objective invites an appealing interpretation: maximizing the first term encourages selecting an $\ba$ that our model is overall uncertain about, and minimizing the second term encourages selecting an $\ba$ for which the individual models in $(\theta_1, \ldots, \theta_N)$ can make confident predictions about the outcome $\feas$.  If we think of the overall uncertainty as a combination of \emph{epistemic} and \emph{aleatoric uncertainty}, then this objective seeks an experiment with high overall uncertainty and low \emph{aleatoric uncertainty}, which therefore has high \emph{epistemic uncertainty}.  Intuitively, if the various possible models in $\Pr(\Theta \mid \Data)$ are individually confident but about differing outcomes, then observing the $\feas$ value corresponding to $\ba$ is likely to prove some of those outcomes incorrect.

Using an ensemble of equally weighted parameter vectors ($\theta_1, \ldots, \theta_N$) to represent $\Pr(\Theta \mid \Data)$ allows us to compute a global feasibility prediction,
\[
\widehat{\Pr}(\Feas = \feas \mid \ba; \Theta) = \frac{1}{N} \sum_{i=1}^N \Pr(\Feas = \feas \mid \ba; \theta_i)
\]
as well as find the experiment that maximizes the estimated \bald{} objective in the form: \begin{equation}\label{eq:final-criterion}
\textsc{bald}(\ba; \Theta) = \ent(\widehat{\Pr}(\Feas \mid \ba; \Theta)) - 
                          \frac{1}{N}\sum_{i=1}^N \ent(\Pr(\Feas \mid \ba; \theta_i)).
                          \end{equation}
                          
\subsection{Sampling Strategies}
\label{sec:sampling-strategies}

Now that we have established an informational score for experiments, we consider several sampling strategies for optimizing over $\A$, the set of plans up to a fixed length $L$.
Maximizing the \bald{} objective over the entire set $\A$ is difficult because we need to consider all discrete plans up to length $L$, as well as all possible assignments to each continuous abstract action parameter. %Additionally, information gain is in general non-convex.

\textbf{Complete} One strategy we consider is uniformly sampling complete plans from $\A$ and scoring the samples. We call this the \complete{} strategy. 
\begin{equation}
\argmax{\ba \in \A}  \text{\bald{}}(\ba ; \Theta_{comp})
\end{equation}

Unfortunately, to achieve a consistent sampling density, the number of required samples scales exponentially with the length of the plan. Additionally, this strategy might generate most of its samples in the infeasible part of the space.

%We observe that the \emph{infeasible sub-sequence property} provides a strong constraint on the set of feasible plans: sequence $\as$ is only feasible if all of its prefixes are also feasible. As such, when calculating the feasibility of a plan, we use our model recursively, to determine the likelihood of success given each subsequence is successful.
%The probability that the plan is feasible is:
% \begin{equation}
%     \Pr(\Phi=\bm1|a_{1:L}, \Theta) = \prod\limits_{l=1}^L \Pr(\Phi=\bm1|a_{1:l}; \Theta),
% \end{equation}

% We observe that the \emph{infeasible sub-sequence property} provides a strong constraint on the set of feasible plans: sequence $\as$ is only feasible if all of its prefixes are also feasible. As such, the model can learn the probability a specific action is feasible instead of the entire plan. If we denote the feasibility of each individual action as $\Phi_l$, then the probability that the plan is feasible is:
% \begin{equation}
%     \Pr(\Phi_{1:L}|\ba, \Theta) = \prod\limits_{l=1}^L \Pr(\Phi_l|a_{1:l}, \Phi_{1:l-1} = \bm1; \Theta),
% \end{equation}
% where $\Phi_l$ represents whether action $a_l$ is feasible and we use the subscript $1:L$ to refer to a sequence of variables.

\textbf{Greedy} Another strategy requiring fewer samples is a \greedy{} approach, in which we select the next action $a_n$ which maximizes the \bald{} objective, given that we have already optimistically constructed $a_{1:n-1}$. This strategy does not take into consideration that if a plan fails early, we do not get to learn from the full plan execution.

We can leverage additional structure afforded to us by the \emph{infeasible subsequence property} when we do active learning using the $\Theta_{ss}$ model class.
% Since this model class is constrained to predict feasibility of an action given the previous actions were feasible, we need a strategy to efficiently sample plans where this property holds. 
In the following, we consider two possible strategies.
%The goal is to select plans where multiple of the action feasibility labels will be informative.

\textbf{Sequential} When generating a potentially informative plan for the $\Theta_{ss}$ model, we can take into account the probability a specific action will be attempted (i.e., that the plan was successful up until that action).
This allows us to find plans whose informative outcomes have a high probability of being observed.
The resulting objective is:
%Fortunately, the $\Theta_{ss}$ model class can be used to predict the probability that we will get to observe the outcome of a specific action:
% To generate an informative plan, we need to take into account the probability a that we get to observe the outcome of each action (i.e., that the plan was successful up until that action). The $\Theta_{ss}$ model class is well suited to this task:
\begin{equation}
\argmax{a_{1:n} \in \A} \sum_{i=2}^n \Pr(\Feas_{1:i-1} = \bm1 \mid a_{1:i-1}; \Theta_{ss}) \text{\bald{}}(a_{1:i} ; \Theta_{ss})
\end{equation}
where the probability is calculated as in Equation \ref{eq:theta-ss}. 
This equation computes the expected information gain for executing a plan, taking into account the probability that plan execution fails at any given step as predicted by the learned \apf{} model.
We refer to this as the \sequential{} strategy.

We implement the \sequential{} approach naively by sampling and scoring entire plans. Like  \complete{}, this method requires exponentially more samples for longer plans, however, in the domain we considered, a sampling approach was acceptable as we had a relatively short plan horizon. In the future work we hope to extend this strategy to a search-based method which prunes candidate plans by their predicted feasibility.

\textbf{Incremental} We also consider a strategy where we only consider plans $\as$ for which we have already observed the prefix to be feasible. In other words, the prefix $(a_1, \ldots, a_{n-1})$ together with result $\feas_{1:n-1} = \bm1$ are in the current data set $\Data$.  We call this the \incremental{} strategy.%We also consider a strategy which uses the \emph{infeasible sub-sequence property} to bias our sampling towards plans that are much more likely to be feasible, and therefore informative. 
\begin{equation}
\argmax{a_{1:n} \in \A} \mathbbm{1}_{(a_{1:n-1}, \feas=1) \in \Data} \text{\bald{}}(a_{1:n} ; \Theta_{ss})
\end{equation}

Although this strategy finds more feasible plans than \complete{}, it also requires the robot to reuse plan prefixes, so we do not gather novel observations when repeating a plan prefix. Constraining our experiments to build on previously feasible plans may be overly restrictive.

\section{Implementation}
\label{sec:implementation}
We have implemented this framework for a class of problems in which the robot manipulates objects to construct towers.  All of our experiments use the $7$-DOF Panda robot from Franka Emika, in simulation and in the real world.

\begin{figure}[t!]
\begin{center}
\includegraphics[width=0.47\linewidth]{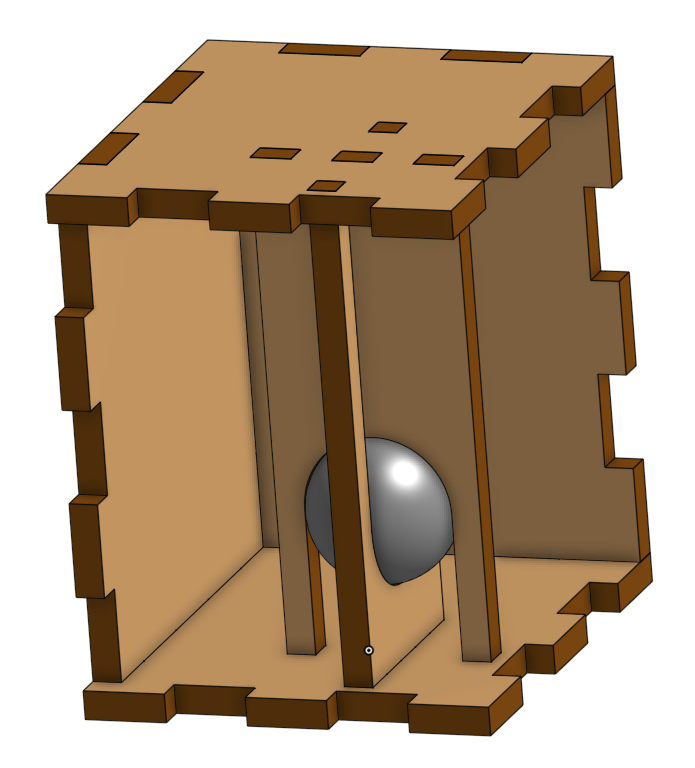}
\includegraphics[width=0.45\linewidth]{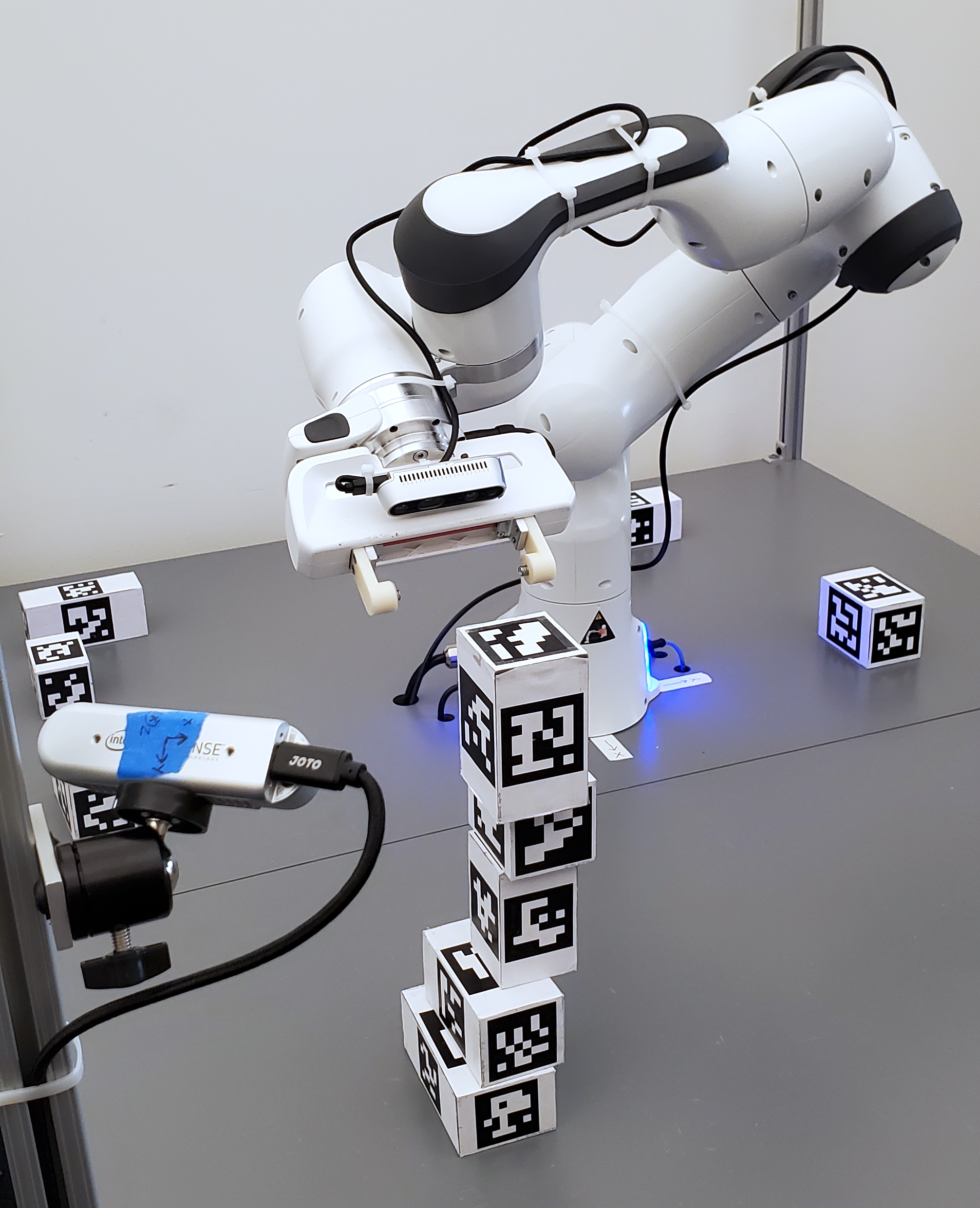}
\end{center}
\caption{\label{fig:panda_and_objects} Left: The cuboids for this manipulation task were constructed from laser cut plywood. A 25mm diameter lead ball is mounted inside some of the objects to significantly alter the mass distribution. Right: Unique ArUco markers are applied to each face of each object, for object identification and localization. Visible in the foreground is one of the two external cameras mounted around the workspace.}
\vspace{-10pt}
\end{figure}

\subsection{Domain}
\label{sec:domain}

% State: Robot configuration, pose, dimensions, CoM for each object.
The world consists of the robot and a set of objects, $\mathcal{O}$, with which it can interact. 
In this work, we consider cuboids with non-uniform mass distributions (Figure \ref{fig:panda_and_objects}).  Each object, $o \in \mathcal{O}$, is described by a tuple, $(d, c, m)$, where $d \in \mathbb{R}^3$ are the dimensions, $c \in \mathbb{R}^3$ is the offset of the center of mass from the center of geometry, and $m \in \mathbb{R}$ is the object's mass.

During the experimentation phase we use a set of 10 blocks, and all evaluations are performed with a different set of 10 blocks. 
The block parameters from each set are sampled from the same uniform distribution over the dimensions of the objects and the locations of the center of mass within the objects.

The abstract actions are to place objects onto a stack;  they are specified by $a=(o, r)$ where $r \in SE(3)$ is the relative pose of $o$ with respect to the object placed in the previous action (or the table if this is the first object placement).

An abstract plan is feasible if the detailed planning and execution system can find and execute robot commands (i.e., grasp poses and motion plans) such that the objects are placed on top of one another and the resulting tower is stable.  Note that, for this property to hold, each prefix of the plan must also be feasible --- that is, each subtower is stable.

The learned \apf{} model is applied to three different objectives in the \emph{evaluation phase}: 
\begin{enumerate}
    \item \emph{Tallest Tower:} The objective is to construct the tallest possible tower.
    \item \emph{Longest Overhang:} The objective is to construct the tower with the maximum distance from the center of geometry of the bottom block to the furthest vertical side of the top block.
    \item \emph{Maximum Unsupported Area:} The objective is to construct the tower where each block has as much area possible unsupported by the block below it.
\end{enumerate}

See Appendix \ref{sec:appendix-gen} for a discussion on the generalizability of our method outside of the towers domain.

\subsection{Perception, planning, and execution}

% While the robot is autonomously learning an \apf{} model and performing tasks, it will generate plans that need to be executed on the real objects. This can require non-trivial motion planning; for example, the current world state might be the result of a previous attempt at building a tower having fallen over, resulting in multiple blocks jumbled on eachother at odd angles. \inote{this paragraph feels like it stops short for some reason}

\textbf{Perception}  
For the system to robustly pick up blocks and recover from unstable towers falling in unpredictable configurations, we require a perception system that can identify and localize objects at arbitrary positions. Although more advanced perception systems might be needed for arbitrary objects, vision is not the immediate focus of this work, so we pattern our objects with ArUco markers to simplify perception. To indicate identity and avoid orientation ambiguity, each object has a unique ArUco marker on each face.

Two RealSense D435 depth cameras mounted statically on a frame observe the workspace and allow for localizing the objects with minimal occlusions. If an object is not visible, it is assumed to be at its home position behind the arm. Due to the resolution of the cameras and size constraints of the tags on the blocks, we found that the pose estimates from the static cameras can have up to 1 centimeter of error. This level of error is acceptable, as rough pose estimates are refined with a third RealSense D435 camera mounted on the robot wrist. As the arm moves to a pre-grasp pose computed from the noisy object pose estimate, the wrist-mounted camera collects images closer to the the object to be grasped, allowing for a refined pose estimate and more precise grasp.

% To grasp an object with pose estimated by the, the robot moves to a pre-grasp pose computed based on th
% Once the robot has moved to an approach location to grasp the object, it uses a wrist camera to fine-tune the object's pose estimate.
% All the cameras used in the work are D435 Realsense cameras, and extrinsic calibration was performed using the MoveIt calibration package.\footnote{\url{https://github.com/ros-planning/moveit\_calibration}} \inote{I think we can skip the camera deets?}.

\textbf{Planning} In this domain, executing an abstract action requires a multi-step task and motion plan. For example, to place an object on a tower with an arbitrary relative pose to the block below it, regrasping may be necessary. When a tower falls over, the robot will also need to move fallen blocks out of the way to build the next tower.

To handle these scenarios, we turn to a large body of work in task and motion planning. We use PDDLStream \cite{garrett2020pddlstream}, which integrates PDDL task-level planning with lower level motion planning. The PDDL domain describes actions including picking and placing objects and moving the arm through free space, while a Bidirectional RRT \cite{kuffner2000rrt} in joint configuration space performs motion planning and collision checking with a surrogate world model implemented in PyBullet \cite{coumans2016pybullet}. In this work we make the simplifying assumption that there are dedicated positions for the base of the tower, regrasping, and storage for each of the objects. 
These constraints are specified to the planner to reduce planning time by limiting the search space of possible action parameters.

\textbf{Execution} The motion plans generated by PDDLStream are executed using joint-space controllers on the robot. When constructing a tower in the experimentation phase, after each block placement the wrist camera is used to check for tower stability. If a tower was unstable, then the last manipulated block will not be near its expected pose in front of the gripper.

To improve data collection efficiency, we parallelize execution and planning. As the robot executes a motion plan to assemble or disassemble a tower, the planner produces plans to move each individual block in that tower under the assumption that all actions will be successful. This parallelism is interrupted if a tower falls over prematurely, prompting a replan to clear the fallen blocks.

If the state of the world is such that a robot is unable to find a plan to proceed with experimentation or execute an existing plan, human intervention may be required. We have provisioned for several of these cases, including blocks falling off the table, or too close to one another for the planner to find feasible grasp candidates. Once such issues are manually resolved (e.g., by putting the block back on the table), the robot can update its estimate of block poses and resume planning.

\subsection{Learning}
\label{sec:imp-learning}
% Details to include:
% - Number of ensemble models
% - NN architecture/hyper-parameters
% - Cross validation set used for early stopping
% - How many samples are used to generate new towers
% - Loss function for an individual model
% - PyTorch implementation
% - Active Learning initialized with 40 random towers.
% - Data augmentation.

\begin{figure}
    \centering{\includegraphics[width=\columnwidth]{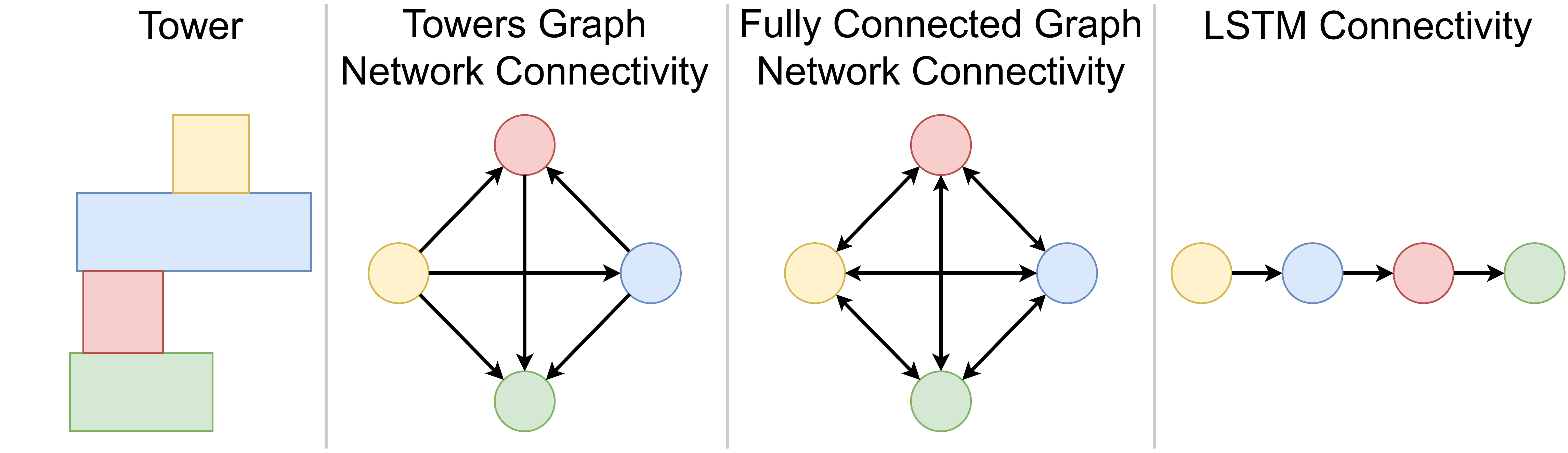}}
    \caption{The flow of information for the model architectures compared in Section \ref{sec:eval-graph}. The \tgn{} uses domain specific connectivity, while the {\sc fcgn} assumes no prior knowledge on how the blocks should be used together to inform predicting feasibility. The {\sc lstm} simply iterates through the blocks starting with the top block.}
    \label{fig:architecture_graphic}
    \vspace{-10pt}
\end{figure}

As discussed in Section \ref{sec:method-apf-model}, the distribution over \apf{} model parameters is represented by an ensemble of networks.
In our implementation, the ensemble is made up of {\sc gnn}s with domain-specific connectivity.
The input to the {\sc gnn} is an abstract action sequence, $\ba$.
Each action, or block placement $a_i$, is passed into a separate node in the graph, and each node (corresponding to a block placement in the tower) is connected to nodes above it in the tower. 
This mirrors the analytical computation of tower stability, in that each subtower's combined center of mass must be within the contact patch of the block below it.
We refer to this network architecture as a \tgn{}, or {\em Towers Graph Network}.
Other network architectures which are invariant to task plan length are a \emph{Fully Connected Graph Network} ({{\sc fcgn}}) and a {\sc lstm} model. 
The {{\sc fcgn}} uses the same node and edge networks as our \tgn{} model, but has edges between each node. 
The {\sc lstm} model passes each block's vector representation through the network in order starting from the top of the tower, and most closely matches our \tgn{} connectivity. 
A visualization of the connectivity of each architecture is given in Figure \ref{fig:architecture_graphic}.

In our experiments, $10$ networks are used in the ensemble.
Each individual network is randomly initialized and trained using the binary cross-entropy loss function with early stopping according to the loss on a validation set, which is also collected actively.

Before active experimentation, each model in the ensemble is initialized by training on the same dataset (shuffled differently for each) of $40$ randomly generated towers. For the  \emph{complete} strategy the towers are of size $2-5$, and for the \emph{sequential}, \emph{incremental}, and \emph{greedy} strategies they are of size $2$.
During the experimentation phase, at each iteration the top $10$ most informative towers are chosen and labeled by attempting to build each with the robot.
$20\%$ of the collected data is added to a validation set and the remainder is added to the training set.
We perform data augmentation by rotating each collected tower $90, 180, \text{and}, 270$ degrees about the axis normal to the table surface.

%\section{Introduction}
%\label{sec:intro}
%\input{sections/intro.tex}

%\section{Method Overview}
%\label{sec:method}
%\input{sections/method.tex}

%\section{Active Learning}
%\label{sec:active}
%\input{sections/active.tex}

%\section{Evaluation Domain}
%\label{sec:domain}
%\input{sections/domain.tex}

\section{Evaluation}
% I think we need to explain what part of x the robot has control over (the pose, not the other object properties)

% Goals of evaluation: (1) Compare different active learning/experiment design strategies in terms of data efficiency. (2) Compare different NN architectures for this problem. (3) Real-robot experiments, ability to collect data on real robot and justify using a learned model.
In the following sections we evaluate the utility of various sampling strategies (Section \ref{sec:eval-sequential}) and model architectures (Section \ref{sec:eval-graph}) for increasing data efficiency and generalization.
% Types of evaluation, regret over time, accuracy over time.
In simulation, we show that these choices greatly influence the robot's accuracy and performance on multiple downstream tasks.
In addition, Section \ref{sec:eval-robot} gives the performance of our \incremental{} sampling strategy on a real robot. We show that not only can the robot learn an \apf{} model from real data and use it to perform downstream tasks, but also that learning on the real robot allows us to be robust to noise that would be difficult to model in simulation.

%Metrics are reported separately for towers of different sizes to highlight the difficulty of larger towers.
%For accuracy evaluations, we evaluate the model's ability to correctly predict plan feasibility using a test-set of novel objects and an equal split between feasible and infeasible plans.
When evaluating task performance, we randomly select $5$ blocks from a set of novel blocks and execute the best tower (that uses all $5$ blocks) found by the Monte Carlo planner, given the task objective and our \emph{Learned} \apf{} model.
The reward received from executing this tower is used to calculate normalized regret, which is the difference between this received reward and the largest reward of a stable tower considered by the planner (found using an \emph{Analytical} model).
If a tower is unstable, we assign a reward of zero.
% NOTE: Mike, removed the below paragraph as it is repeated later in the real robot section.
%In our real robot experiments we also compare performance when using different \apf{} models: an \emph{Analytical} model assumes there is no execution noise in placing blocks, and a \emph{Simulation} noisy model samples $10$ towers with noisy block placements from a given normal distribution and considers a tower feasible if all sampled towers are analytically stable.
%In Section \ref{sec:eval-sequential} we show performance on the three different downstream tasks given in Section \ref{sec:tasks}, and highlight the importance of feasible action data acquisition.
%In Section \ref{sec:eval-graph} we highlight the benefits of our graphical model in capturing block stacking dynamics, and how it enables generalization outside of our training data.
%We also show the distribution of the acquired data, and how our graphical model enables us to use this data effectively.
%Finally, in Section \ref{sec:eval-robot}, we perform real robot learning, planning, and execution to show the importance of learning with real-world data as compared to a simulated and analytical models.

\subsection{Impact of Sampling Strategy}
\label{sec:eval-sequential}

\begin{figure*}
    \centering
    \begin{subfigure}{0.32\textwidth}
        \centering
        \includegraphics[width=\textwidth, trim={.5cm 0 1.5cm 0}, clip]{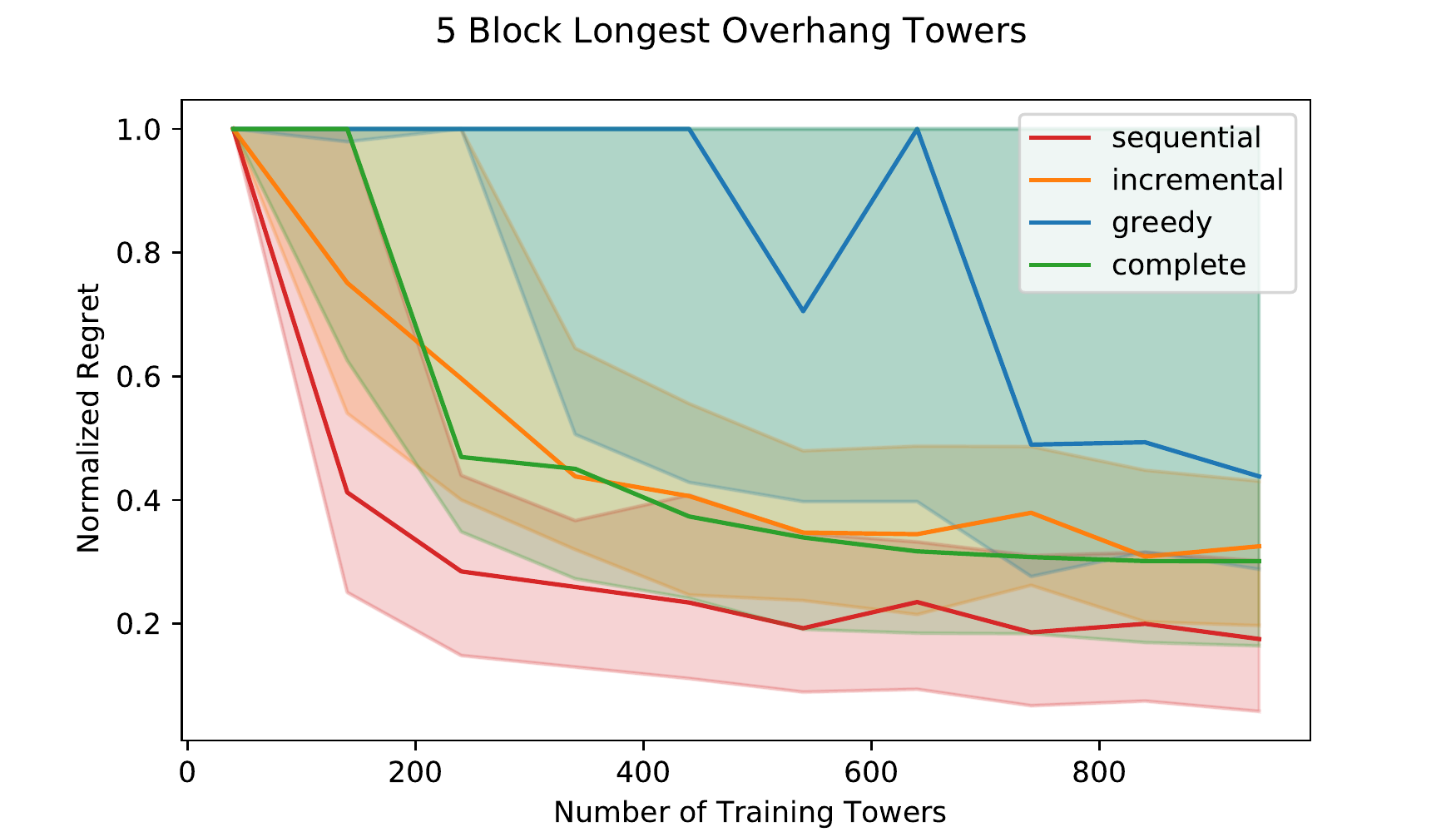}
        \label{fig:eval-strategies-overhang}
    \end{subfigure}
    \begin{subfigure}{0.32\textwidth}
        \centering
        \includegraphics[width=\textwidth, trim={.5cm 0 1.5cm 0}, clip]{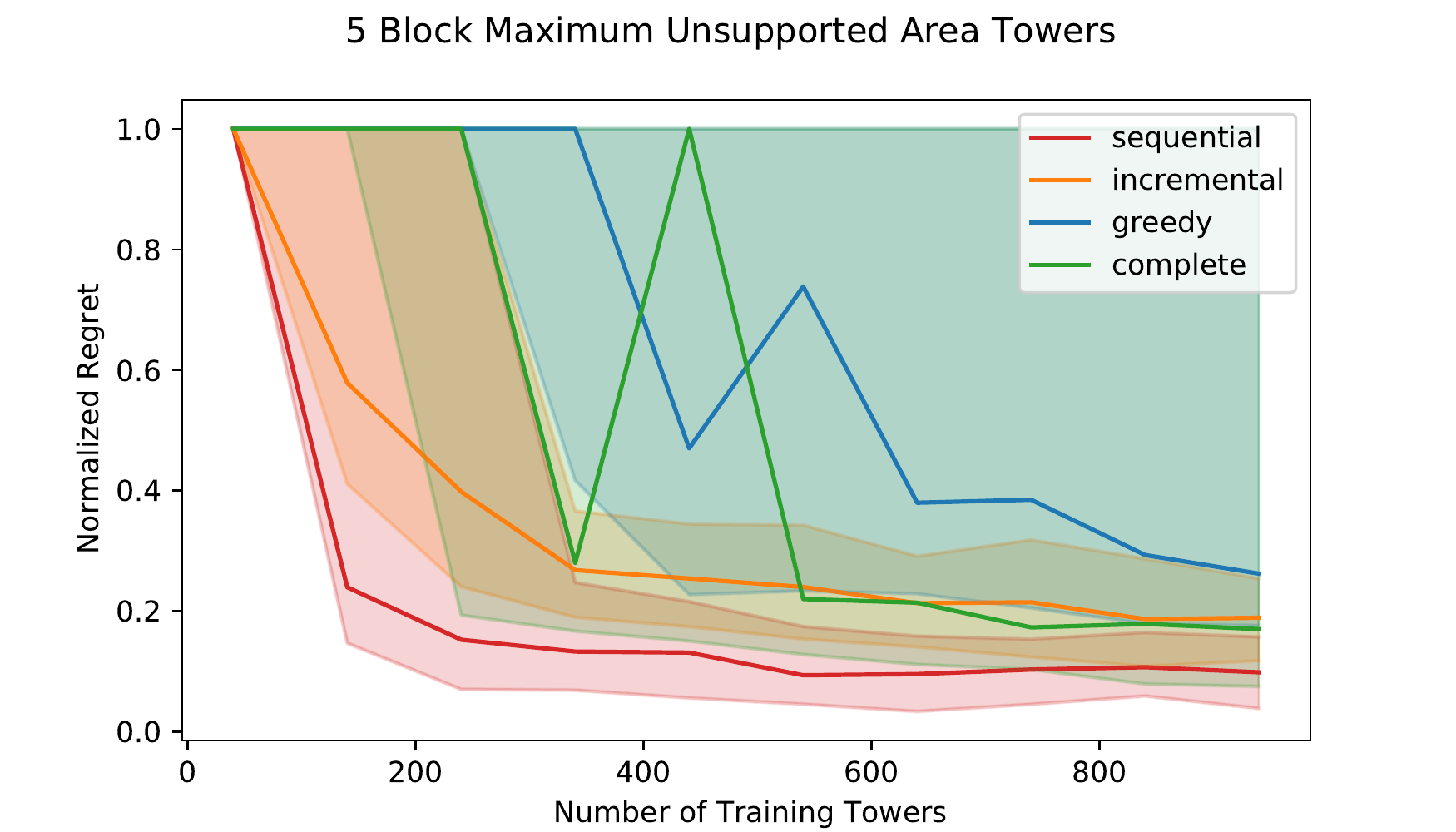}
        \label{fig:eval-strategies-contact}
    \end{subfigure}
    \begin{subfigure}{0.32\textwidth}
        \centering
        \includegraphics[width=\textwidth, trim={.5cm 0 1.5cm 0}, clip]{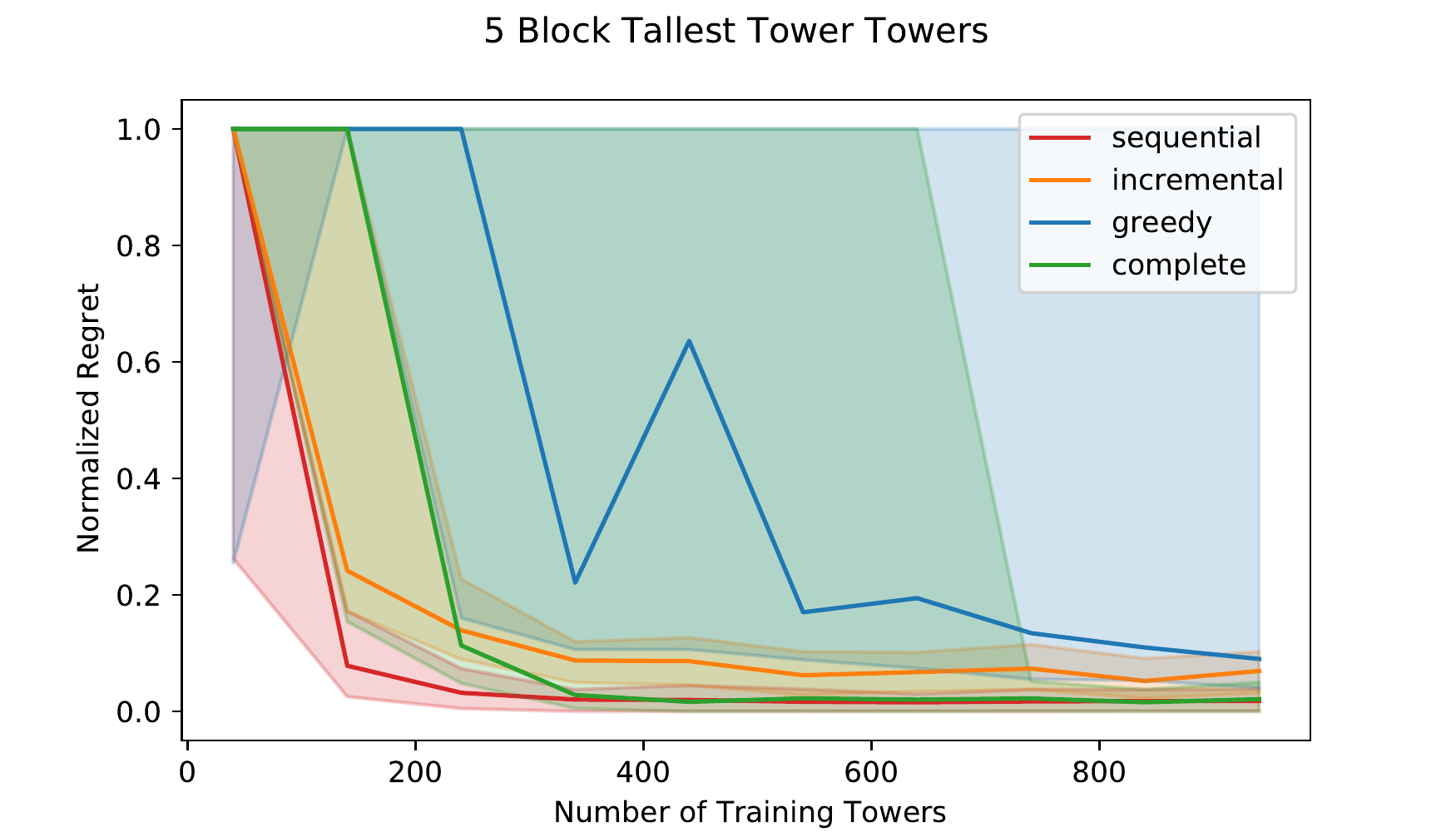}
    \label{fig:eval-strategies-tallest}
    \end{subfigure}
        \vspace{-10pt}
        \caption{A comparison of sampling strategies on different downstream tasks all performed in simulation. Each method evaluation consists of $4$ separate \apf{} model-learning runs, and each point is the Median Normalized Regret of $50$ individual planning runs per learned model. The shaded regions show $25\%$ and $75\%$ quantiles.}
        \label{fig:eval-strategies}
        \vspace{-12pt}
\end{figure*}

Figure \ref{fig:eval-strategies} shows the task performance of models trained using the four sampling strategies described in Section \ref{sec:sampling-strategies}.
Each strategy has results aggregated from $4$ independent training runs and $50$ task evaluations per run all performed in a simulated environment. Different sets of blocks are used between training and evaluation. 

Our \apf{} model performs best on the \emph{Tallest Tower} task, successfully minimizing regret after constructing only $200$ towers with the \sequential{} method. 
The \incremental{} method also performs well, successfully minimizing regret with minimal variance across runs. The \complete{} method on average performs well, but has very high variance for the more challenging tasks, \emph{Longest Overhang} and \emph{Maximum Unsupported Area}. Note that the shaded region represents the quartile distribution --- a region that extends to $1.0$ means that more than a quarter of the trials were unstable. This highlights the importance of considering the \emph{infeasible sub-sequence property} when sampling and scoring plans in the action space. 
Finally, the naive \greedy{} strategy performs the worst, and is only able to achieve decent performance on the \emph{Tallest Tower} task after seeing roughly $800$ training towers, likely due to the fact that it is not considering the feasibility of subtowers when searching the actions space, just greedy single-block placements.

The \emph{Maximum Unsupported Area} and \emph{Longest Overhang} tasks are more challenging for the robot because they require deep understanding of the tower stability decision boundary, while the \emph{Tallest Tower} task only requires a rough understanding of how to build stable towers with high confidence. These results show that in spite of the difficulty of the first two tasks, the active learner is able to improve its understanding of the decision boundary well enough to perform tasks with very low regret and low variance.

In Appendix \ref{sec:appendix-sim-random} we give additional results which compare against baselines that do not leverage active learning.

% Discuss the benefits of using our specific GNN structure compared to other alternatives (in particular, LSTM and fully connected GNN).

\subsection{Model Architecture and Generalization}
\label{sec:eval-graph}

\begin{figure}
    % \vspace{-5pt}
    \centering{\includegraphics[width=\columnwidth]{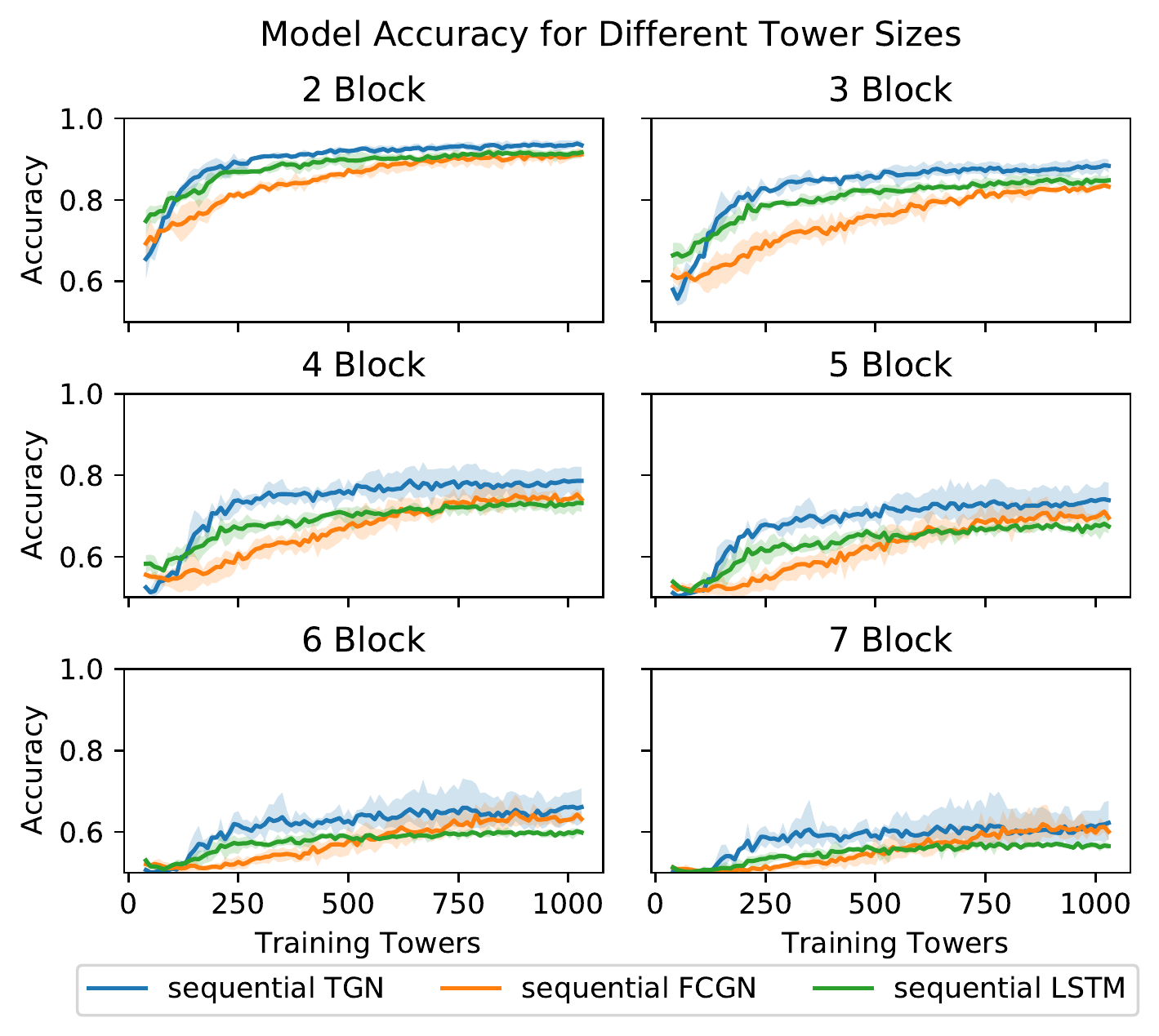}}
    \caption{A comparison of different network architectures using the \sequential{} sampling strategy, performed in simulation. The accuracy for each method is averaged over 3 separate training runs. The shading shows the minimum and maximum accuracy from these runs.}
    \label{fig:eval-models}
    \vspace{-12pt}
\end{figure}

We compare our \tgn{} to the other network architectures discussed in Section \ref{sec:imp-learning} and shown in Figure \ref{fig:architecture_graphic}.
For this evaluation, we report accuracy on a held-out test set of towers built with a novel set of blocks.
The test set consists of half feasible and half infeasible towers, and $1000$ towers for each tower size.
Our models were trained in simulation on towers consisting of up to $5$ blocks, but our results give model accuracy for towers ranging from $2$ to $7$ blocks, shown in Figure \ref{fig:eval-models}.

%While this seems unintuitive based on common {\sc lstm} approaches which take in actions in order of execution, this matches more closely the implementation of our \tgn{} model which is modelled after the analytical calculation of stability starting from the top block.
%system using several models for learning Abstract Plan Feasibility. We consider Recurrent Neural Networks (specifically LSTMs) as well as Graph Neural Networks (GNNs), and we compare both to a baseline collection of Multi-Layer Perceptrons (MLPs) for fixed plan-lengths. Both the LSTM and GNN models can generalize to plans of arbitrary length -- in this domain towers of arbitrary height -- while the MLP models cannot.

%In the towers domain where plan feasibility is the ability to construct a tower without it falling over at any point, it is necessary to consider the relative poses of all the blocks in the tower. 
In the towers domain, it is necessary to consider the joint centers of mass for groups of blocks above support blocks. 
While the {\sc lstm} architecture could remember the previous blocks as it iterates through the tower, in practice we find that it is outperformed by the graph network architectures. 
We believe this is because the connectivity of the graph networks allows them to precisely compare adjacent blocks in addition to aggregating information about multiple blocks. 
The weakness of the {\sc lstm} is more pronounced as the number of blocks in a tower increases.

Our \tgn{} architecture is structured to be biased towards our particular domain, so it is able to improve its predictions much faster than the other architectures.
This enables good planning time performance as seen in Section \ref{sec:eval-sequential} with similar long-term performance to the {\sc fcgn} architecture.

% Interestingly, the {\sc fcgn} has better long-term performance.
% We believe that not biasing the structure of the graph network towards a particular domain makes for a more challenging learning problem initially, but in the long run the true underlying function is able to be learned more precisely.
% However, for the purpose of learning on a real robot platform, good planning performance with fewer samples is preferable.
%We experiment with fully connected GNNs, as well as introducing a domain-inspired sparsity pattern into the graph which restricts nodes to only receive information from the nodes corresponding to blocks physically above them in the tower. 

\subsection{Real Robot Experiments}
\label{sec:eval-robot}

Finally, we give results for executing the entire active learning pipeline on a real Panda robot (see Section \ref{sec:implementation} for details of the real robot setup).
In total, the robot built $400$ towers while training over a period of $55$ hours.

For the experimentation phase, we used a fixed set of $10$ training blocks. The \tgn{} ensemble is initialized with $40$ random $2$-block towers labeled in simulation with added relative-pose noise. We generated candidate experiments using the \incremental{} strategy described in Section \ref{sec:sampling-strategies}, and produced stability labels for the constructed towers by observing the outcome with the cameras.

%\footnote{We were limited in our access to the robot due to Covid-19 restrictions and could not repeat this experiment for the other sampling strategies.}

During the evaluation phase, we test the robot's ability to use its learned \apf{} model to perform all tasks described in Section~\ref{sec:domain} with a separate set of $10$ held-out evaluation blocks. 
We compare the learned \apf{} model to two baselines.
First, we compare to a hand-engineered model of plan feasibility that calculates whether a candidate tower is feasible in a noiseless world.
We also compare to a noisy simulator feasibility model, which predicts a tower is feasible only if the candidate tower is also stable to $10$ normally distributed perturbations, with a standard deviation of $5$mm, for each block placement.

% \begin{table*}[!h]
% \centering
% \footnotesize
% \begin{tabular}{|c||c|c|c||c|c|c||c|c|c|}
% \hline
%  & \multicolumn{3}{c||}{\textbf{Longest Overhang}} & \multicolumn{3}{c||}{\textbf{Max Unsupported Area}} & \multicolumn{3}{c|}{\textbf{Tallest Tower}}\\ 
% \hline
% \textbf{APF Model} & \textbf{Regret} & \textbf{Stable Regret} & \textbf{\# Stable}& \textbf{Regret} & \textbf{Stable Regret} & \textbf{\# Stable}& \textbf{Regret} & \textbf{Stable Regret} & \textbf{\# Stable}\\ \hline
% Learned Panda Model & 0.45 & 0.38 & 9/10 & 0.33 & 0.25 & 9/10 & 0.15 & 0.05 & 9/10\\
% \hline
% Simulation (5mm noise) & 0.47 & 0.33 & 8/10 & 0.19 & 0.19 & 15/15 & 0.41 & 0.01 & 6/10\\
% \hline
% Analytical & 0.80 & 0.00 & 2/10 & 0.80 & 0.00 & 2/10 & 0.40 & 0.00 & 6/10\\
% \hline
% \end{tabular}
% \caption{\label{tab:real-results}
% Real robot task performance when using different \emph{Abstract Plan Feasibility} models. The \emph{Learned} model was trained with data collected through active learning on the real Panda robot. The \emph{Analytical} and \emph{Simulation} models calculate feasibility using the known underlying dynamics but use either no noise or simple noise models respectively.} 
% \end{table*}

\begin{table}[b]
\centering
\scriptsize
\begin{tabular}{|c||c|c||c|c||c|c|}
\hline
 & \multicolumn{2}{c||}{\textbf{Tallest}}
 & \multicolumn{2}{c||}{\textbf{Longest}} 
 & \multicolumn{2}{c|} {\textbf{Max Unsupported}} \\
 & \multicolumn{2}{c||}{\textbf{Tower}}
 & \multicolumn{2}{c||}{\textbf{Overhang}} 
 & \multicolumn{2}{c|} {\textbf{Area}} \\
\hline
\textbf{Model} & \textbf{Regret} & \textbf{\#Stable}& \textbf{Regret} & \textbf{\#Stable}& \textbf{Regret} & \textbf{\#Stable}\\ \hline
Analytical & 0.30 & 7/10 & 0.80 & 2/10 & 0.80 & 2/10\\
\hline
Simulation & 0.41 & 6/10 & 0.47 & 8/10 & 0.19 & 10/10\\
\hline
Learned & 0.15 & 9/10 & 0.45 & 9/10 & 0.33 & 9/10\\
\hline
\end{tabular}
\caption{\label{tab:real-results}
Real robot task performance when using different \apf{} models. The \emph{Learned} model was trained with data collected through active learning on the real Panda robot. 
The \emph{Analytical} and \emph{Simulation} models calculate feasibility using the known underlying dynamics but use either no noise or simple noise models respectively.} 
\vspace{-3pt}
\end{table}

\begin{figure}[t!]
    \centering{\includegraphics[width=0.48\textwidth,trim=0 0 900px 0, clip]{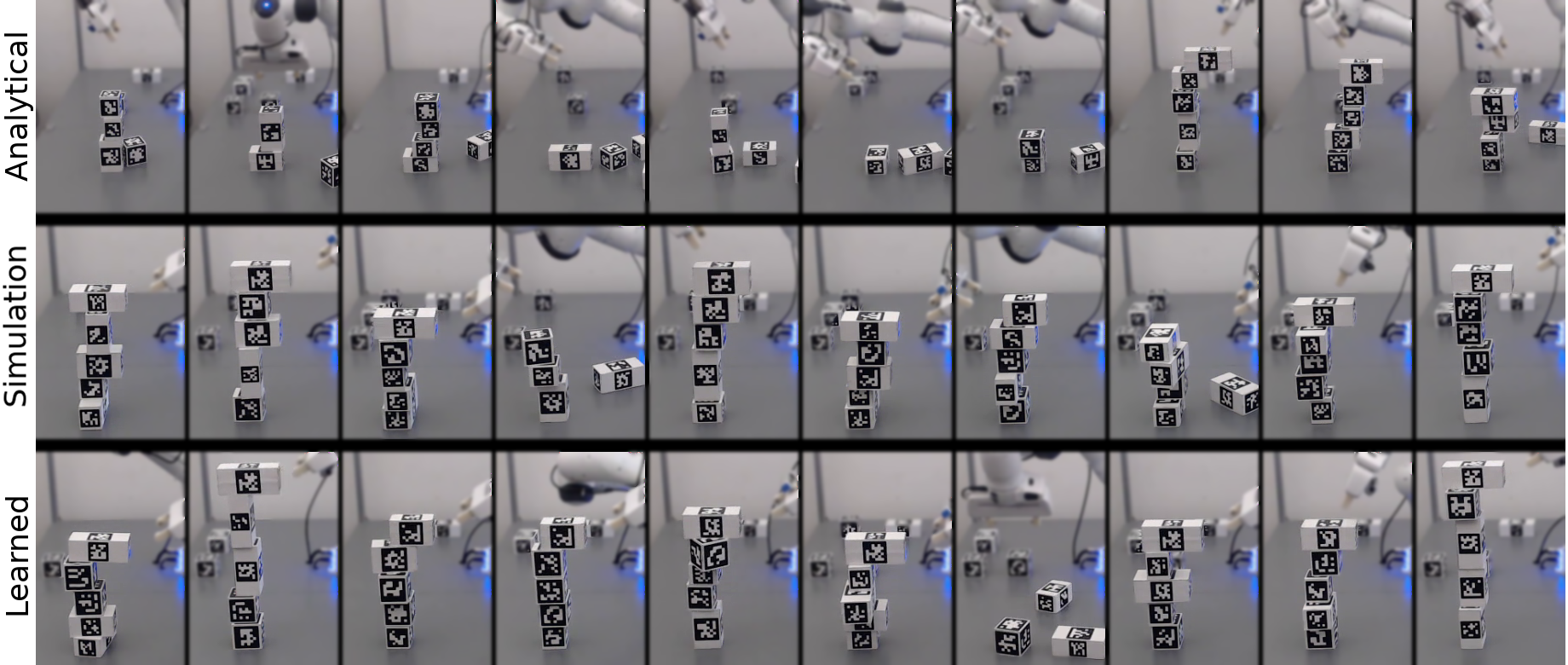}}
    \caption{Towers built for the \emph{Longest Overhang} task when using different \apf{} models. Observe that some towers built using the \emph{Analytical} and \emph{Simulation} models were unstable. See the appendix for more examples.}
    \label{fig:panda_eval_qualitative}
    \vspace{-10pt}
\end{figure}

For each task and model we select $5$ blocks at random from the evaluation set and plan to maximize the given objective.
If a constructed tower is unstable, it gets zero reward.
The robot constructs $10$ towers for each task and feasibility model.
In Table~\ref{tab:real-results}, we report average normalized regret, and the number of total trials that resulted in a stable tower.
% We also report stable regret --- regret averaged only across the towers that were stable.
Figure~\ref{fig:panda_eval_qualitative} includes images of the robot performing the \emph{Longest Overhang} task using the different models for \apf{}.

From these results, it can be seen that the \emph{Analytical} model can build towers with high reward when the tower is stable, but the towers it chooses to build are rarely stable across all three tasks. 
However, the \emph{Simulation} and \emph{Learned} \apf{} models can still build towers with large overhang while considering the effects of noisy action execution on a real robot.
% While our \emph{Learned} model exhibits slightly higher stable regret, it outperforms the \emph{Simulation} model in stability of constructed towers, and therefore results in overall lower regret when taking stability into account.
% This is due to the fact that the \emph{Simulation} model makes assumptions about the type of noise distribution (Gaussian only in the plane normal to the table) and its parameters (variance).
% The \emph{Learned} model, one other hand, learns a more complex noise model that more accurately reflects the real world.
Our \emph{Learned} model performs competitively with the \emph{Simulation} model, and in aggregate leads to similar stability across all tasks (27 versus 24 stable towers out of 30).
However, note that the \emph{Simulation} model presents higher variability in tower stability across tasks.
This is because the model makes assumptions about the type of noise distribution (Gaussian only in the plane normal to the table) and its parameters (mean and variance), which may be more suitable for certain tasks.
The \emph{Learned} model, on the other hand, may capture other complex real-world phenomena that significantly contribute to plan feasibility in a task-agnostic setting.

% Although the \emph{Analytical} model can achieve perfect regret when towers are stable, the towers it chooses to build are rarely stable across all three tasks.
% This is because it does not take into account noise that is present when executing on a real robot.
% The \emph{Simulation} model that includes simulated noise, outperforms the \emph{Analytical} model in terms of stability.
% The \emph{Learned} \apf{} model is able to achieve high reward while consistently building more stable towers across all tasks.

%As training progresses, the robot consistently builds more stable towers that have high reward as dictated by the respective task objectives.
% \textcolor{red}{SC: This will now be a figure comparing the towers over different constructability models, right?}

% \begin{figure}[!h]
%     \centering{\includegraphics[width=\columnwidth]{figures/qualitative.png}}
%     \caption{Qualitative results showing task success vs. number of data acquisition steps.}
%     \label{fig:panda_eval_qualitative}
% \end{figure}

% To demonstrate the effectiveness of our active learning system, we have summarized some basic timing and robustness statistics from our experiments in Table \ref{tab:panda_stats}.

% \begin{table}[!h]
% \caption{\textbf{Robot Experiment Statistics}}
% \centering
% \label{tab:panda_stats}
% \begin{tabular}{|c|c|}
% \hline
% Total training time & 55 hrs \\
% \hline
% Average pick-and-place time & 300 s \\
% \hline
% Number of total pick and place tasks & 1200  \\
% \hline
% Number of manual interventions & 125 \\
% \hline
% Number of errors requiring restart & 50 \\
% \hline
% Block pose estimation error & 2 cm \\
% \hline
% Placement error & 5 mm \\
% \hline
% \end{tabular}
% \end{table}

\section{Related Work}
\label{sec:related_work}
% 1) Section on related planning work that considers feasibility.
% tabouda2016coarse_feasibility: Take advantage of a high-level abstract plan before doing low-level motion planning.
% singh1996feasible: use feasible action constraints to avoid planning in configuration space for tasks with complex dynamics and obstacles.
% kim20a: Learning a value function from solving task instances in geometric domains. Value function is estimated using a GNN and used to guide search on more difficult problems.
% dreiss20 (deep visual heuristics): Predict feasibility of discrete actions in sensor space. Used a passive simulated dataset with 20k scenes to predict feasibility of grasps.
% wells2019: Learn a feasibility model to predict the feasibility of a motion plan. Used to address long timeouts given by infeasible plans when using a probabilistic motion planner in the inner loop.
% toussaint_multi: Multi-level logic and geometric solvers.
% (not currently discussed) A Long Horizon Planning Framework for Manipulating Rigid Pointcloud Objects 

\textbf{Hierarchical Planning} 
It is a common approach to long-horizon planning problems; decomposing the solution into high level reasoning over \emph{abstract actions} and lower level reasoning over \emph{concrete actions} \cite{lozano-perez_constraint-based_2014, tabouda2016coarse_feasibility, toussaint_multi-bound_2017}.
\citet{singh1996feasible} performed early work on considering the feasibility of high level plans to improve efficiency when planning in high dimensional spaces.
%In order to constrain the search space of the planning problem, \citet{singh1996feasible} propose to only consider actions that are inherently feasible.
Recent works have proposed methods that predict feasibility of an action as a way to reduce the number of calls to expensive solvers and enable more efficient planning \cite{driess2020deep, wells2019learning}.
Our work builds upon this literature by presenting a method to learn feasibility actively when detailed physical models are not available.

%\citet{tabouda2016coarse_feasibility} leverage \emph{Abstract Plan Feasibility} for motion planning in reach and avoid problems. They obtain a coarse-model with an obstacle-informed discretization of the workspace, which allows them to discard infeasible high-level plans and lazily defer lower-level motion planning. 
%Much earlier, \citet{singh1996feasible} use feasible action constraints to avoid planning in configuration space for tasks with complex dynamics and obstacles. Their approach instead plans in \textit{action space}, where they optimize over a set of short-horizon plans constrained by their feasibility under the model. In contrast to both of these approaches, we opt to learn an abstract plan feasibility model. This obviates the need to engineer a model, and also enables feasibility planning in domains which are otherwise poorly modeled.

\textbf{Learning for Task and Motion Planning}
In this work we leverage a task and motion planning framework to plan for concrete actions, then execute those actions to determine plan feasibility.
Others have explored learning the feasibility of actions, specifically an action's preconditions and effects.
\cite{wang2018active} uses a Gaussian process to learn these action parameters, with a specialized acquisition function to guide learning. 
Our work differs in that our acquisition strategy considers observations of entire plans, which means our active learning is non-myopic. 
In addition, for the stacking domain we found graph neural networks to be a more applicable class of function approximators in which plan length can vary.
A similar approach is taken in \cite{kaelbling2017learning}, a precursor to \cite{wang2018active}.

% 2) Active learning.
% MacKay 1992: Early info theoretic active-learning.
% Houslby 2011: Extension to infinite dimensional parameters.
% Gal, 2017: Deep neural nets.
% Beluch, 2018: Use ensembles!

\textbf{Active Learning} Active learning \cite{mackay1992objective} is a well-established learning paradigm that aims to minimize the number of samples needed to learn the target concept.
Recently, \citet{gal_deep_2017} have extended {\sc bald} \cite{houlsby_bayesian_2011} to complex model domains of deep neural networks using MC-dropout \cite{gal2016dropout}.
However, in this work, we follow the approach of \citet{beluch_power_2018} and use an ensemble of deep networks for active learning.

%  Curiosity-driven Exploration by Self-supervised Prediction:
% - active-learning/curiosity (BALD, deepak pathak)
% pathak2019: Self-Supervised Exploration via Disagreement: Use disagreement in an ensemble to learn a forward dyanmics model. They only consider short horizon tasks.

Ideas from active learning have been used for efficient learning in model-based reinforcement learning tasks \cite{pathak2019self,shyam2019model}.
\citet{pathak2019self} explore the environment by taking actions that maximize disagreement between an ensemble of forward models. 
Instead of predicting continuous states, our work learns a model that predicts feasibility of \emph{abstract plans} where the focus is on performing long-horizon tasks.

% 3) Talk about work that predicts compositional dyanmics.

% - Simulation as an engine of physical scene understanding
% Xia: Learning relational transition models.
% - A Compositional Object-Based Approach to Learning Physical Dynamics
% - Relational inductive bias for physical construction in humans and machines
% graph networks (for physical reasoning)/model-based learning
% - Interaction Networks for Learning about Objects, Relations and Physics
% Summary: Learns a dynamics function by factoring the dynamics into object-level dynamics and interaction terms and structuring a NN according to these relations. The relations in this work are known, unlike in the NPE
% - Graph Networks as Learnable Physics Engines for Inference and Control

\textbf{Learning Dynamics} Many recent works have focused on learning predictive dynamics models in scenes with varying numbers of objects.
In such scenarios, it has been shown that explicitly representing objects and their relations in the model can lead to more efficient learning and generalization \cite{battaglia2013simulation, battaglia2016interaction, chang2016compositional, xia2018learning}.
As such, graph networks are becoming a more common modeling choice in these domains \cite{kim2020learning, sanchez2018graph}.
Specifically, in a stacking domain, \citet{hamrick_relational_2018} show that a graph neural network can predict stability properties of towers when given access to a large training set.

% 4) Stacking/Blocks world domain.
%\subsection{Stacking Task}
% Main differnce is that we learn feasibility from an actively gathered dataset that
% takes into account robot errors.
\textbf{Stacking Domain}
Tasks that involve stacking objects in a \emph{Blocks World} have a long history in artificial intelligence. 
Early works developed methods to compute the stability of block placements and construct a target configuration \cite{blum1970stability, winston1972}.
More recently, computer vision researchers have developed scene understanding algorithms that take into account known geometries and stability properties of objects within the scene \cite{GuptaEfrosHebert_ECCV10, jia_3d_2015, roberts1963machine}.
\citet{furrer2017autonomous} developed a system that can build stacks out of stones using detailed models of the objects.

Recent work has shown the ability to predict tower stability using deep learning techniques \cite{ferrari_shapestacks_2018, hamrick_relational_2018, lerer2016learning}.
However, typically these works have used passively collected datasets which include orders of magnitude more samples than required in this work --- making them infeasible to actively collect on a real robot.
An active learning approach allows the robot to explore efficiently, and learn a feasibility model under the real-world noise distribution.
In addition, vision-based systems would not be effective when the state contains non-visual properties, such as the center of mass in our stacking domain \cite{ferrari_shapestacks_2018, lerer2016learning}. 
% stacking task
% - ShapeStacks: Learning Vision-Based Physical Intuition for Generalised Object Stacking.
% - Summary: A dataset of stability prediction for stacking. 20000 examples. Also consider building stacks in sim without a real robot.
% - Blocks World Revisited: Image Understanding Using Qualitative Geometry and Mechanics (added)
% - 3D-Based Reasoning with Blocks, Support, and Stability (jia_3d_2015) added
% - Autonomous Robotic Stone Stacking with Online next Best Object Target Pose Planning
% - Learning to Exploit Stability for 3D Scene Parsing

% Numerous works have learned to predict the stability of towers based on images \cite{lerer2016learning, li2016fall, li2017visual}. Although these approaches can learn to outperform humans, they typically require hundreds of thousands to millions of training examples, necessitating the use of simulated data. They also struggle to generalize to new objects and new scenes whose visual appearance differs from those on which they were trained.

\section{Conclusion} 
\label{sec:conclusion}
We have presented a system which leverages information-theoretic active learning to acquire an \emph{Abstract Plan Feasibility} model, and shown that incorporating plan feasibility into the active learning strategy can dramatically improve sample efficiency. 
We deployed our system on a real Franka Emika Panda robot arm in a block-stacking domain, enabling the robot to learn a useful \apf{} model with only $400$ experiments.

% To apply our method to a new domain, the overall methodology would remain unchanged, but the key components will need to be adapted to that domain.
% Specifically, one would need a) a definition of \emph{abstract actions} and their corresponding parameters, b) \emph{experimental infrastructure} to autonomously execute abstract actions in the physical world, c) an \apf{} detector to provide labels during experimentation and evaluation, and d) (optional) \emph{inductive bias} in a function approximator (e.g., our {\sc tgn} architecture) to further improve data efficiency. 

In future work, we are eager to apply this approach to other domains.
We believe that this self-supervised method of curious exploration is an exciting direction, as it may someday allow the millions of robots sitting powered-off in laboratories around the world to make effective use of their downtime.

\section*{Acknowledgments}
\label{sec:acknowledgements}
The authors would like to thank Rachel Holladay and Caelan Garrett for contributing their time to help us set up our Panda robot arm and get our footing using the Franka Emika and PDDLStream software. This research was generously sponsored by Honda Research Institute.

%% Use plainnat to work nicely with natbib. 

\bibliographystyle{plainnat}
%\bibliography{references}

\clearpage

\appendix
\subsection{Real Robot Evaluations}
\label{sec:appendix-real-robot}

\begin{figure*}[t!]
    \centering{\includegraphics[width=0.85\linewidth]{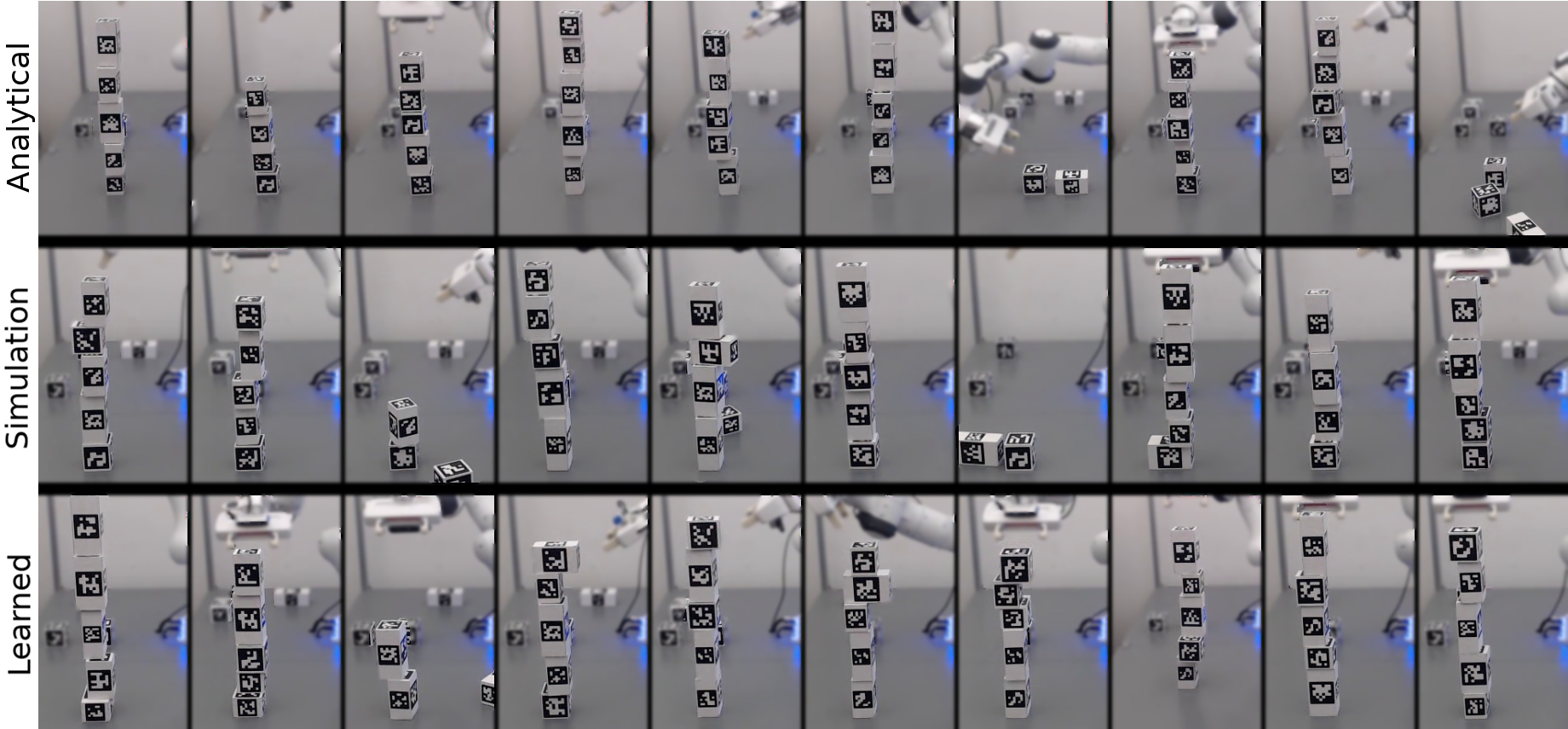}}
    \caption{Towers built for the \emph{Tallest} task when using different \emph{Abstract Plan Feasibility} models.}
    \label{fig:towers-tallest}
\end{figure*}

\begin{figure*}[t!]
    \centering{\includegraphics[width=0.85\linewidth]{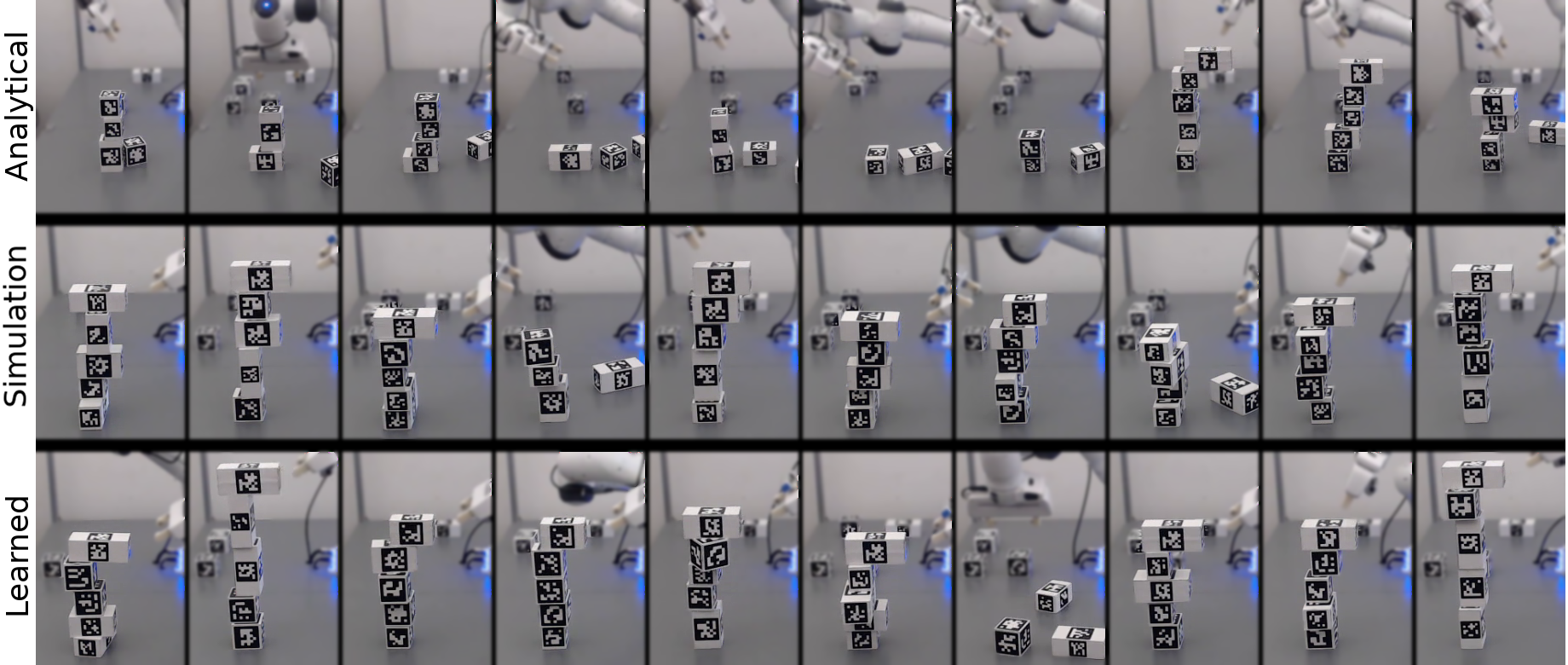}}
    \caption{Towers built for the \emph{Longest Overhang} task when using different \emph{Abstract Plan Feasibility} models.}
    \label{fig:towers-overhang}
\end{figure*}

\begin{figure*}[t!]
    \centering{\includegraphics[width=0.85\linewidth]{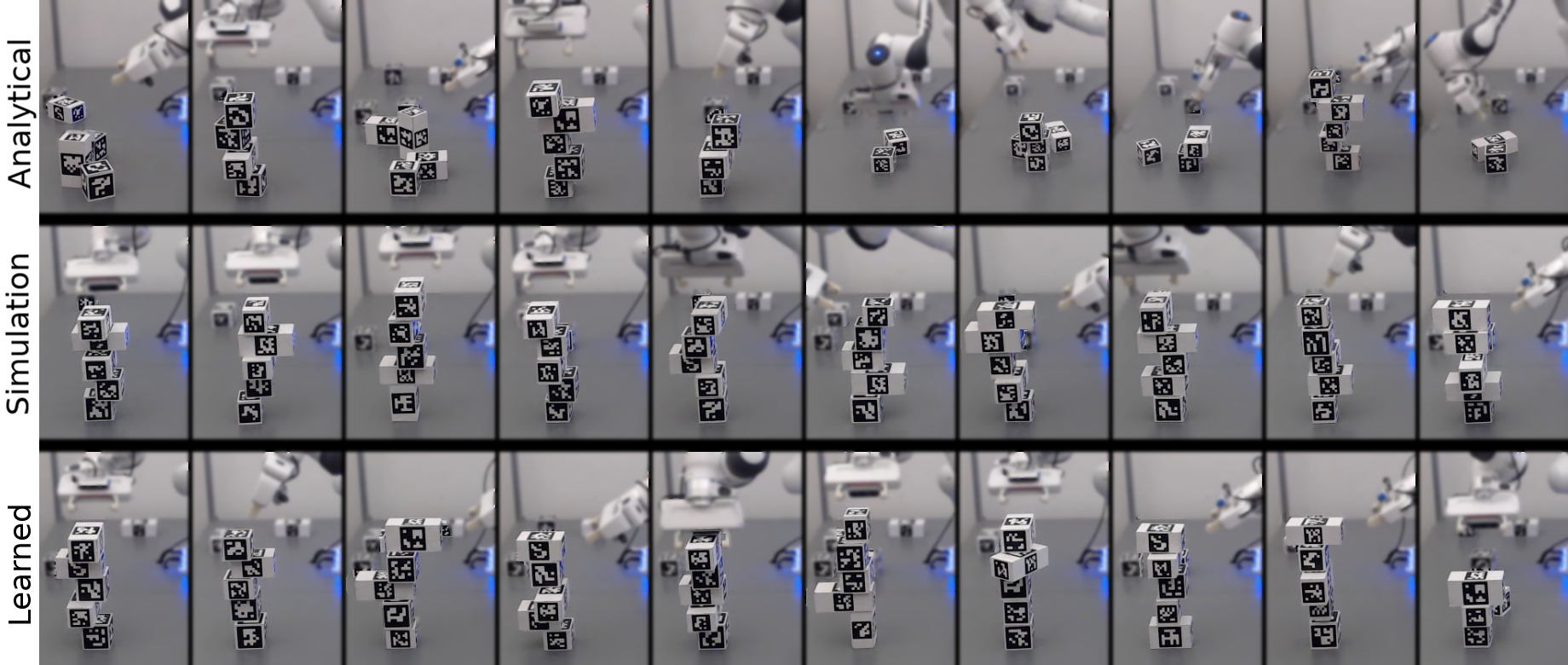}}
    \caption{Towers built for the \emph{Maximum Unsupported Area} task when using different \emph{Abstract Plan Feasibility} models.}
    \label{fig:towers-unsupported}
\end{figure*}

In this section we provide additional results for the experiments described in Section \ref{sec:eval-robot}.
Figure \ref{fig:towers-tallest} shows the towers built when performing the \emph{Tallest Tower} task with each of the analytical, simulation, and learned models.
$10$ towers were built for each task using $5$ blocks.
Towers that have fallen over or do not have $5$ blocks were infeasible plans.
Figures \ref{fig:towers-overhang} and \ref{fig:towers-unsupported} show the evaluation towers for the \emph{Longest Overhang} and \emph{Maximum Unsupported Area} tasks, respectively.

\subsection{Supervised vs Active Learning Comparisons}
\label{sec:appendix-sim-random}

\begin{figure*}
    \centering
    \begin{subfigure}{0.32\textwidth}
        \centering
        \includegraphics[width=\textwidth, trim={.5cm 0 1.5cm 0}, clip]{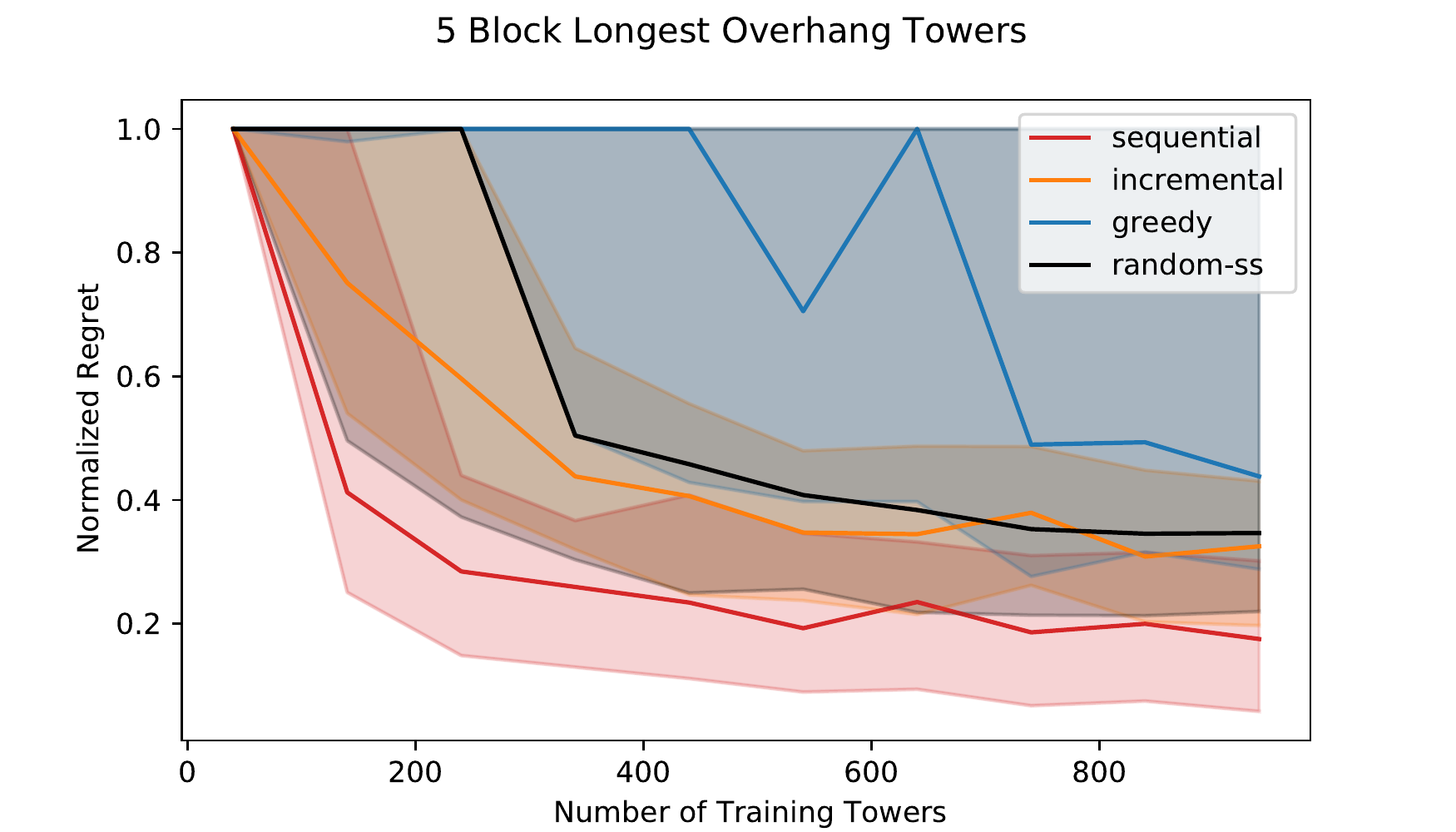}
        \caption{}
        \label{fig:eval-strategies-overhang-ss}
    \end{subfigure}
    \begin{subfigure}{0.32\textwidth}
        \centering
        \includegraphics[width=\textwidth, trim={.5cm 0 1.5cm 0}, clip]{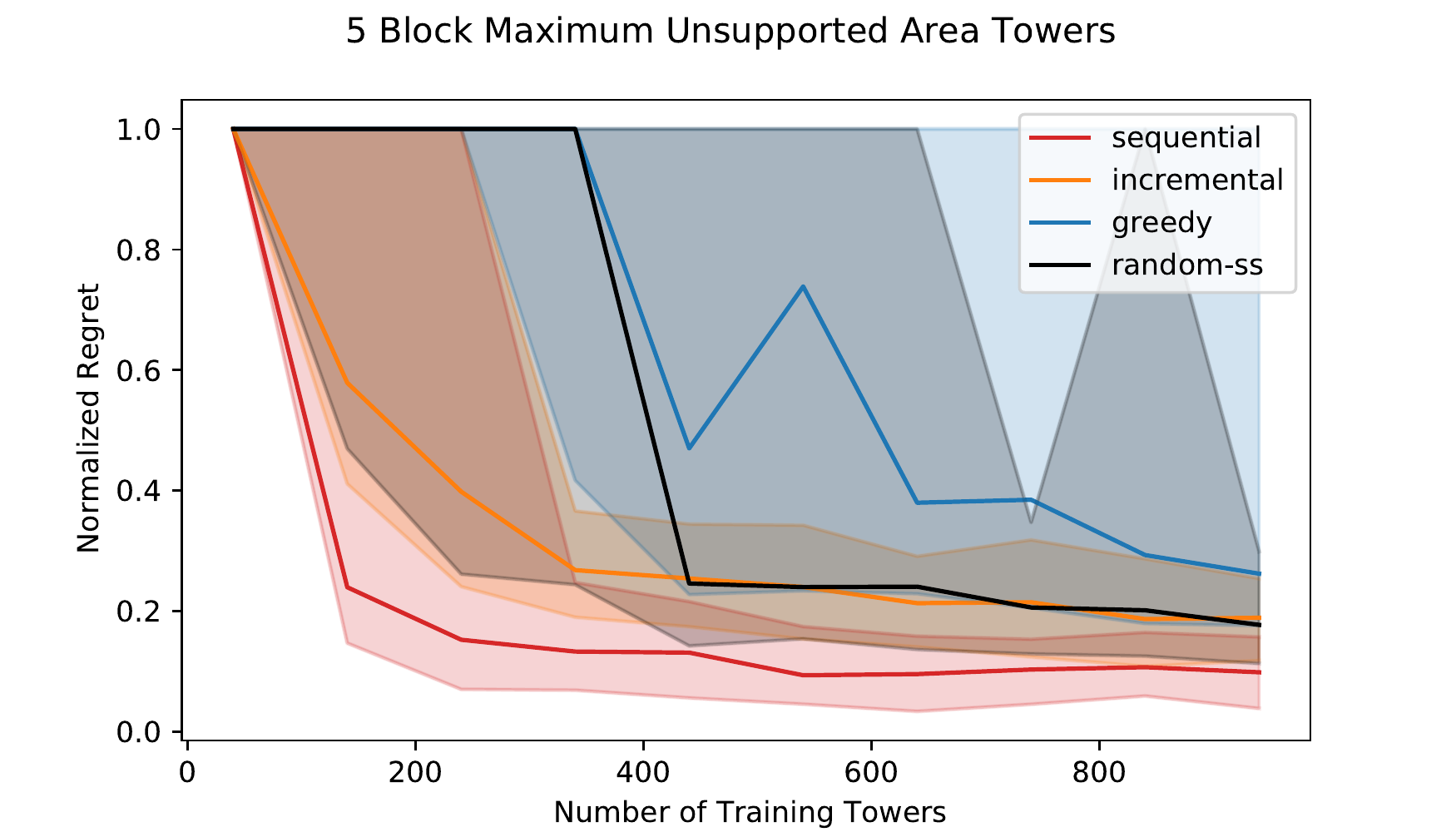}
        \label{fig:eval-strategies-contact-ss}
    \end{subfigure}
    \begin{subfigure}{0.32\textwidth}
        \centering
        \includegraphics[width=\textwidth, trim={.5cm 0 1.5cm 0}, clip]{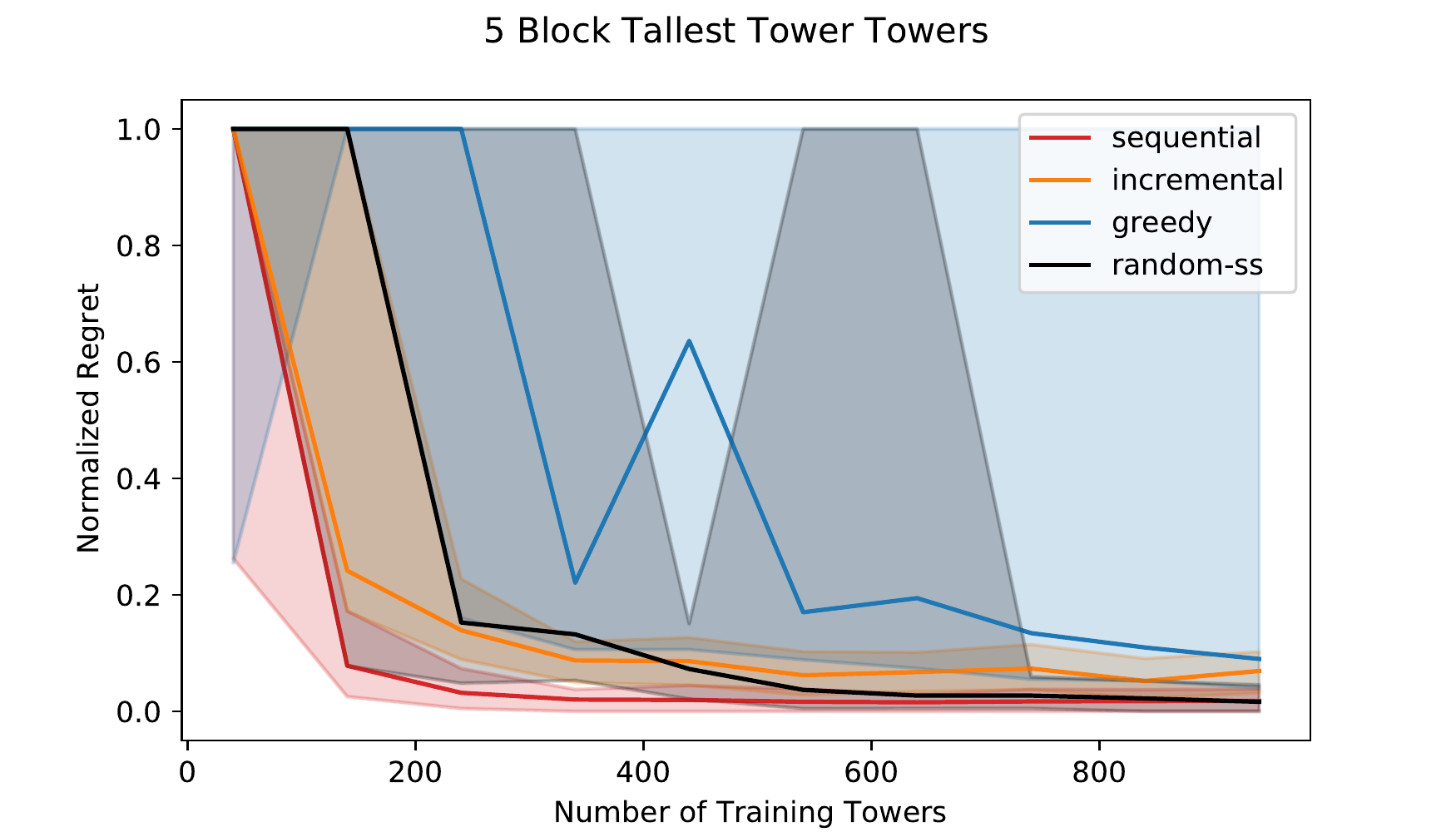}
    \label{fig:eval-strategies-tallest-ss}
    \end{subfigure}
        \caption{A comparison of sampling strategies which use the $\Theta_{ss}$ model class on different downstream tasks all performed in simulation. 
        The \emph{random-ss} strategy randomly samples action sequences and thus does not use active learning.
        Each method evaluation consists of $4$ separate \apf{} model-learning runs, and each point is the Median Normalized Regret of $50$ individual planning runs per learned model. The shaded regions show $25\%$ and $75\%$ quantiles.}
        \label{fig:eval-strategies-ss}
        \vspace{-10pt}
\end{figure*}

In this section we compare the active learning methods described in this paper to supervised learning baselines which do not perform active data collection to train the \apf{} model.
Figure \ref{fig:eval-strategies-ss} compares our methods which use the $\Theta_{ss}$ model class to a \emph{random-ss} strategy. 
\emph{random-ss} randomly samples action sequences and labels each subsequence. The results show that our active methods which leverage the \emph{infeasible subsequence property}, \sequential{} and \incremental{}, outperform \emph{random-ss}. \emph{random-ss} only outperforms \greedy{}, showing that a random strategy is able to find more interesting training towers than a myopic approach which simply tries to maximize the \bald{} objective for a single block placement.

\begin{figure*}
    \centering
    \begin{subfigure}{0.32\textwidth}
        \centering
        \includegraphics[width=\textwidth, trim={.5cm 0 1.5cm 0}, clip]{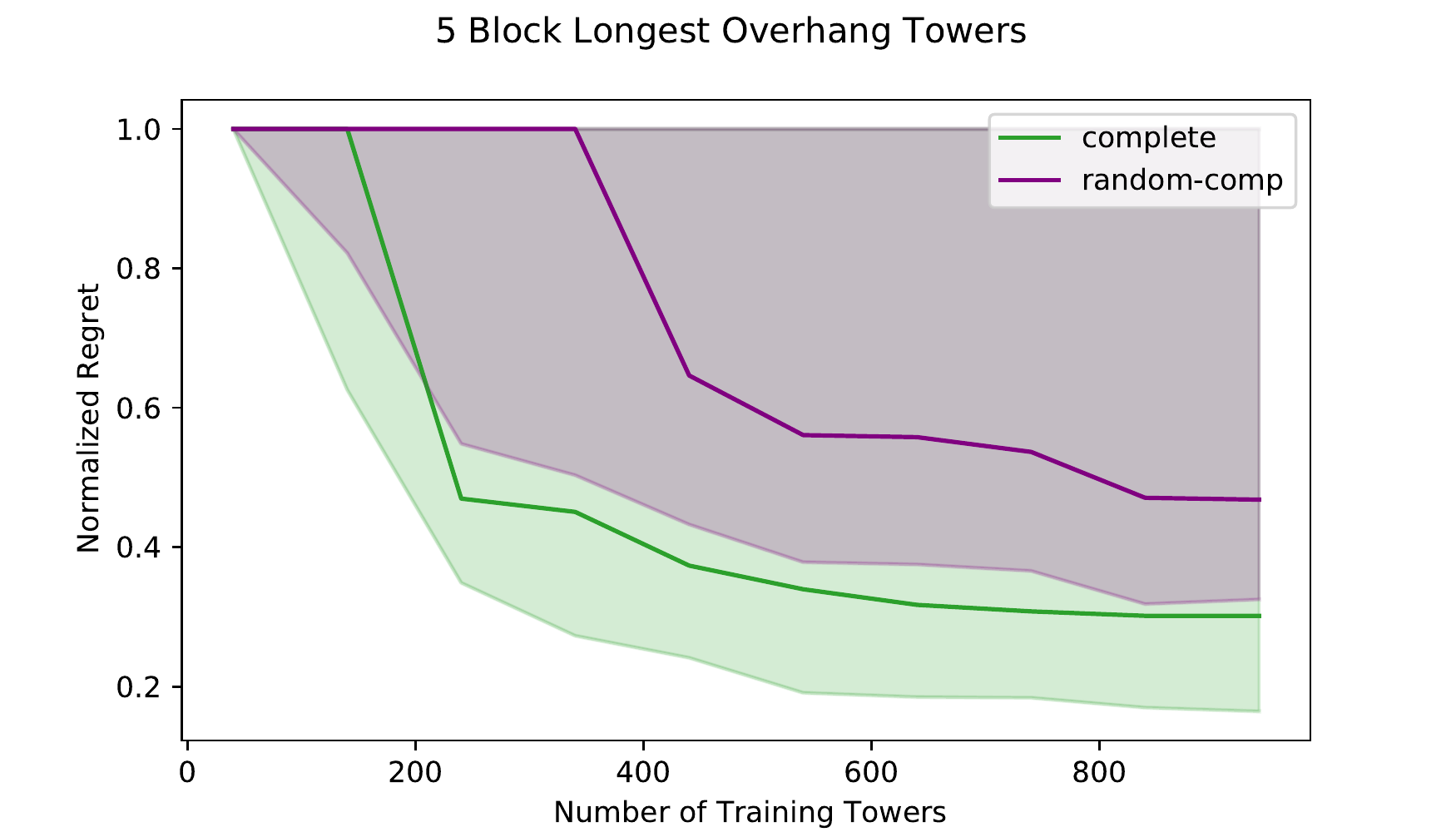}
        \label{fig:eval-strategies-overhang-comp}
    \end{subfigure}
    \begin{subfigure}{0.32\textwidth}
        \centering
        \includegraphics[width=\textwidth, trim={.5cm 0 1.5cm 0}, clip]{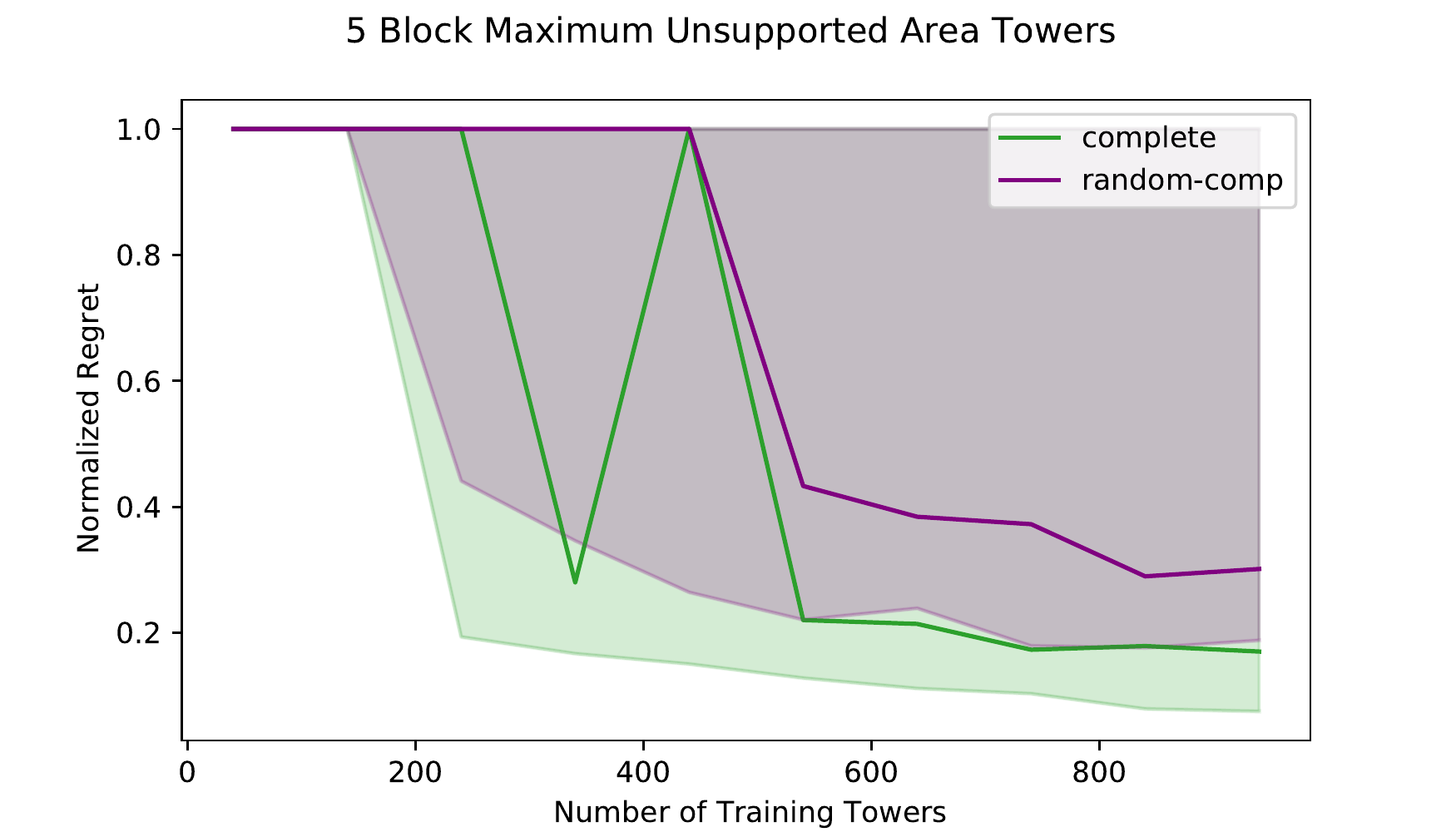}
        \label{fig:eval-strategies-contact-comp}
    \end{subfigure}
    \begin{subfigure}{0.32\textwidth}
        \centering
        \includegraphics[width=\textwidth, trim={.5cm 0 1.5cm 0}, clip]{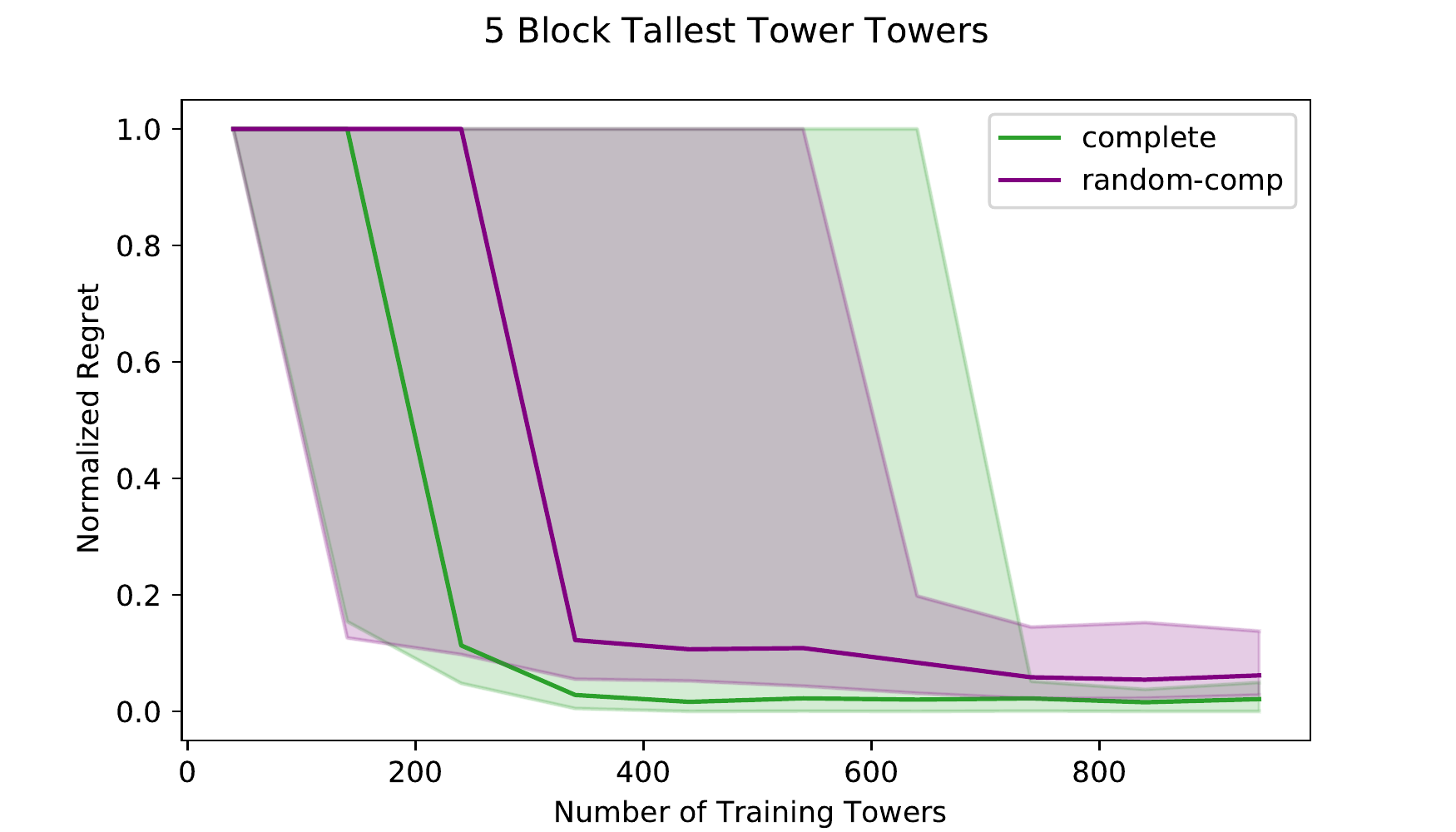}
    \label{fig:eval-strategies-tallest-comp}
    \end{subfigure}
        \caption{A comparison of sampling strategies which use the $\Theta_{comp}$ model class on different downstream tasks all performed in simulation. 
        The \emph{random-comp} strategy randomly samples action sequences and thus does not use active learning.
        Each method evaluation consists of $4$ separate \apf{} model-learning runs, and each point is the Median Normalized Regret of $50$ individual planning runs per learned model. The shaded regions show $25\%$ and $75\%$ quantiles.}
        \label{fig:eval-strategies-comp}
        \vspace{-10pt}
\end{figure*}

Figure \ref{fig:eval-strategies-comp} compares our method, \complete{}, which uses the $\Theta_{comp}$ model class, to a \emph{random-comp} strategy. \emph{random-comp} randomly samples action sequences to train on and only labels the full sequences. As expected, training on sequences which maximize the \bald{} objective is more effective than randomly sampling action sequences.

The data used to train the \emph{random} methods is not actively collected, but we show how training on increasingly larger datasets compares to the actively collected dataset. 

\subsection{Method Generalizability}
\label{sec:appendix-gen}

Here, we motivate a more general class of problems for which our system applies and clarify which components of the system are specific to our chosen domain. 
Our method most benefits domains where the feasibility of an action depends strongly on the preceding action sequence (e.g., adding a fifth block to a tower that already has four blocks). 
Domains that include construction tasks, like ours, will commonly benefit from non-myopic information gathering and a feasibility predictor that incorporates previous actions into its predictions.
To apply our method to a new domain (e.g., consider a packing problem where many objects need to be placed in a larger container), the overall system/methodology would remain unchanged. 
However, one would need to adapt the following domain-specific components:

\begin{enumerate}
    \item \emph{Abstract action definitions}. Parameterize an abstract action that is appropriate for the domain and connect it to concrete actions (e.g., object locations within the container).
    \item \emph{Experimental infrastructure}. Additional capabilities to autonomously plan and execute abstract actions in the physical world (e.g., motion and grasp planning infrastructure).
    %Additional planning capabilities beyond the existing general motion and grasp planning infrastructure, if required.
    \item \emph{Feasibility detector}. The notion of \apf{} will depend on the desired outcome of an abstract action. For a new action, we require a method to autonomously acquire the feasibility label during execution (e.g., whether the gripper will collide when placing the object or if an object will be damaged).
    \item \emph{Additional inductive bias} (optional). Additional structure to the learner can further increase data efficiency, as shown by our {\sc tgn} method. However, this is not required as we show in Figure \ref{fig:architecture_graphic} that a general purpose graph network can be used (e.g., a graph network that has connectivity between all objects).
\end{enumerate}

\end{document}